\documentclass[runningheads]{llncs}
\usepackage{graphicx}

\usepackage{tikz}
\usepackage{comment}
\usepackage{amsmath,amssymb} 
\usepackage{color}
\usepackage{amsmath,graphicx}
\usepackage{booktabs}
\usepackage{multirow}
\usepackage{caption}
\usepackage{subcaption}
\usepackage{amsmath}
\usepackage{amssymb}
\usepackage{comment}
\usepackage{float}
\usepackage{booktabs}
\usepackage{cite}
\usepackage[T1]{fontenc}
\usepackage{textcomp}
\usepackage{soul}
\usepackage{scalerel}
\usepackage{nicefrac}
\usepackage{multirow}
\usepackage{graphicx}
\usepackage{tabularx}
\usepackage{tablefootnote}
\usepackage{verbatim}

\begin{document}
\setcounter{secnumdepth}{3}

\pagestyle{headings}
\mainmatter
\def\ECCVSubNumber{100}  

\title{Monitoring of Pigmented Skin Lesions Using \\ 3D Whole Body Imaging} 

\titlerunning{Monitoring of Pigmented Skin Lesions Using 3D Whole Body Imaging}

\author{David Ahmedt-Aristizabal\inst{1} \and
Chuong Nguyen\inst{1} \and
Lachlan Tychsen-Smith\inst{1} \and
Ashley Stacey\inst{3} \and
Shenghong Li\inst{1} \and
Joseph Pathikulangara\inst{2} \and  \\ 
Lars Petersson\inst{1} \and
Dadong Wang\inst{1}
}
\authorrunning{Ahmedt-Aristizabal et al.}
%
\institute{Imaging and Computer Vision Group, CSIRO Data61, Australia 
\email{\{david.ahmedtaristizabal,chuong.nguyen,lachlan.tychsen-smith, \\ 
shenghong.li,lars.petersson,dadong.wang\}@data61.csiro.au}
\and
Astronomy and Space Science, CSIRO S\&A, Australia \\
\email{joseph.pathikulangara@csiro.au} 
\and
Engineering and Design, CSIRO Data61, Australia\\
\email{ashley.stacey@data61.csiro.au} 
}
\maketitle

\begin{abstract}
\vspace{-2pt}
Advanced artificial intelligence and machine learning have great potential to redefine how skin lesions are detected, mapped, tracked and documented.
Here, We propose a 3D whole-body imaging system known as 3DSkin-mapper to enable automated detection, evaluation and mapping of skin lesions. 
%
A modular camera rig arranged in a cylindrical configuration was designed to automatically capture images of the entire skin surface of a subject synchronously from multiple angles. Based on the images, we developed algorithms for 3D model reconstruction, data processing and skin lesion detection and tracking based on deep convolutional neural networks. We also introduced a customised, user-friendly, and adaptable interface that enables individuals to interactively visualise, manipulate, and annotate the images. The interface includes built-in features such as mapping 2D skin lesions onto the corresponding 3D model.
%
The proposed system is developed for skin lesion screening, the focus of this paper is to introduce the system instead of clinical study. Using synthetic and real images we demonstrate the effectiveness of the proposed system by providing multiple views of a target skin lesion, enabling further 3D geometry analysis and longitudinal tracking.
Skin lesions are identified as outliers which deserve more attention from a skin cancer physician. Our detector leverages expert annotated labels to learn representations of skin lesions, while capturing the effects of anatomical variability. 
It takes only a few seconds to capture the entire skin surface, and about half an hour to process and analyse the images.
%
Our experiments show that the proposed system allow fast and easy whole body 3D imaging. It can be used by dermatological clinics to conduct skin screening, detect and track skin lesions over time, identify suspicious lesions, and document pigmented lesions. The system can potentially save clinicians time and effort significantly.
The 3D imaging and analysis has the potential to change the paradigm of whole body photography with many applications in skin diseases, including inflammatory and pigmentary disorders. 
With reduced time requirements for recording and documenting high-quality skin information, doctors could spend more time providing better-quality treatment based on more detailed and accurate information.

\vspace{-5pt}
\end{abstract}

\section{Introduction}

Regular examination and skin monitoring are essential for the early detection of cutaneous malignancies including melanoma. Monitoring of pigmented skin lesions over a long period of time is often a manual, time-consuming and error-prone task, especially when a patient has hundreds of skin lesions, both the inspection and documentation of these lesions are highly laborious and inefficient~\cite{vestergaard2008dermoscopy}.

Data-driven approaches using dermoscopic images have been developed in recent decades to assist dermatologists in clinical decisions and to spot extremely suspicious situations~\cite{barata2018survey}.
Recent studies have demonstrated the ability of machine learning to match, if not outperform, clinicians in the diagnosis of individual skin lesions in controlled reader studies where dermoscopic images contain a single skin lesion. For example, algorithms derived from the ISIC Grand Challenge~\cite{rotemberg2021patient} have shown the potential to outperform over 500 clinical experts in a controlled environment~\cite{tschandl2019diagnostic}. However, these studies do not reflect clinical scenarios where clinicians need to examine all skin lesions of a patient regularly.

Despite the potential for using deep learning in clinical dermatology such as skin cancer risk assessment~\cite{de2019development}, most systems assume that all relevant lesions of the patient have been manually identified prior to classification. Thus, the detection of skin lesions in wide-field images (photographs depicting multiple lesions from large body parts) is needed~\cite{birkenfeld2020computer}.

Contrary to classic investigations, where the attention is focused on a particular lesion and a single image is taken using a dermoscope or a high-resolution camera, whole body imaging systems capture imagery of the entire epidermis. Such a system enables not only lesion detection, but also the possibility of a longitudinal study of dermatological patients. In this manner, it is possible to detect lesions that evolve over time and to find new lesions that emerge after the previous examination. 

Traditionally, to monitor all skin lesions on the skin surface of a patient, 2D whole body imaging has been employed. The use of 2D whole body imaging has shown the potential to reduce the number of biopsies taken, and increase the accuracy of diagnosis for patients at high risk of melanoma~\cite{truong2016reduction}.
Recent advances in artificial intelligence have demonstrated its capability to support physicians in detecting outliers or suspicious ``ugly duckling lesions'' in wide-field images~\cite{soenksen2021using,mohseni2021can}. A few commercial systems such as Dermengine~\cite{dermengine} support whole body image analysis using mobile cameras.
Nevertheless, the data collection is time and resource intensive and requires a photographer to take photos of subjects in different poses. Thus, skin lesion comparison remains challenging due to changes in the pose or illumination between different scans. The majority of previous registration techniques are evaluated only on a small part of the body (\textit{e.g.} the back or front torso). 

Whole body photography increases the area captured in an image by imaging the majority or all of the skin. Korotkov et al.~\cite{korotkov2018improved} developed a system for acquiring overlapping images of the entire body's skin to facilitate whole body scanning. Lesion tracking across multiple scans was automated, but the system was not extended to include 3D reconstruction of skin surfaces. A device used for capturing images of the whole area of human skin with one camera was also introduced by Strzelecki et al.~\cite{strzelecki2021skin}, but the entire body was not captured simultaneously and tracking of skin lesions across 3D meshes were not explored.
Investigations toward the identification of lesions that differ from most of the other marks on a patient's skin using wide-field images have gained attention for quick assessment of high-risk patients for further examination~\cite{birkenfeld2020computer,soenksen2021using}.
Nevertheless, such correlation methods that identify particular lesions on the patient's body that appear in images taken from different views and times are not well studied in the literature.

\paragraph{Related work: 3D whole body imaging for skin data management} \hfill

Advanced 3D imaging technology, supported by automated analysis methods, is changing the way we diagnose skin malignancies by enabling standardised and gapless image acquisition~\cite{rayner2018clinical}.
3D imaging has enabled domain researchers to gain new understandings at a faster pace from the image data. 3D data exploration enables natural representation and accurate position measurements when particular attributes of interests cannot be easily identified in a 2D image~\cite{treleaven20073d,ahmedt2019vision}. Thus, 3D mapping methods are being used to accurately inspect the fine topology and textual features of a lesion~\cite{rayner2018clinical,bogo2014automated,grochulska2021additive}.

Digital 3D whole body images enable a complete and automatic documentation of all skin lesions of a patient, the determination of their geometric parameters, and their further analysis by a dermatologist~\cite{rayner2018clinical}. 3D imaging allows for 360-degree rotation to view the body from all angles, including the ability to view curved surfaces which are difficult to see in 2D imaging. This technology can also be used to provide context to single lesion dermoscopy images~\cite{janda2021describing}, facilitate long-term surveillance, and support the evaluation of raised lesions, as the 3D images can be viewed from multiple angles. This means that lesions are considered within the context of the surrounding skin, offering a more accurate overall assessment and diagnosis than dermoscopy images alone~\cite{grochulska2021additive}.

\begin{table}[t!]
\centering
\vspace{-9pt}
\caption{Comparison of our proposed system for data management with other related works in 3D whole body imaging.}
\resizebox{0.8\textwidth}{!}{%
\begin{tabular}{
l c c c c c c
}
\toprule
& ~\cite{bogo2014automated} & ~\cite{zhao2021detection} & ~\cite{navarrete2020total} & ~\cite{betz2021reproducible} & ~\cite{grochulska2021additive} & ours
\\
\midrule
Low-cost acquisition system                &            &            &            &            &            & \checkmark \\
2D whole body photography                  & \checkmark &            & \checkmark & \checkmark & \checkmark & \checkmark \\
3D human reconstruction                    & \checkmark &            & \checkmark & \checkmark & \checkmark & \checkmark \\ 
Clinical images used for lesion detection  &            &            & \checkmark & \checkmark & \checkmark & \checkmark \\ 
Data-driven skin lesion detector           &            & \checkmark & \checkmark & \checkmark & \checkmark & \checkmark \\ 
Map lesions to 3D body texture             & \checkmark & \checkmark & \checkmark & \checkmark & \checkmark & \checkmark \\ 
Data-driven 3D meshes registration          &            & \checkmark & \checkmark & \checkmark & \checkmark & \checkmark \\ 
Longitudinal tracking of skin lesions      & \checkmark & \checkmark & \checkmark & \checkmark & \checkmark & \checkmark \\ 
User interface for data management         &            &            & \checkmark & \checkmark & \checkmark & \checkmark \\ 
\bottomrule
\end{tabular}}
\label{table:literature-summary}
\end{table}

The main specifications of the proposed system compared to related works are listed in Table~\ref{table:literature-summary}.
Preliminary 3D body scanning for the detection of new melanocytic lesions is described by Bogo et al.~\cite{bogo2014automated}. Such a solution is based on a multi-camera 3D stereo system to capture body shape and skin texture, and a 3D body model to register scans across time and track lesions based on the registered body locations. However, the model is based on expensive scanning devices and is limited to higher-resolution RGB imagery. The detection of lesions is based on handcrafted features.
Recently, Zhao et al.~\cite{zhao2021detection} used an annotated publicly available 3D textured mesh dataset for detecting and tracking longitudinal skin lesions. Lesions are identified first on the skin using 2D texture images, and then are mapped to their corresponding positions on the 3D mesh surface of the subject as proposed in our system. However, the development of a complete pipeline for capturing whole body images and constructing a patient avatar along with integrated existing lesions is not considered.

\begin{figure}[t!]
\begin{center}
\includegraphics[width=0.99\linewidth]{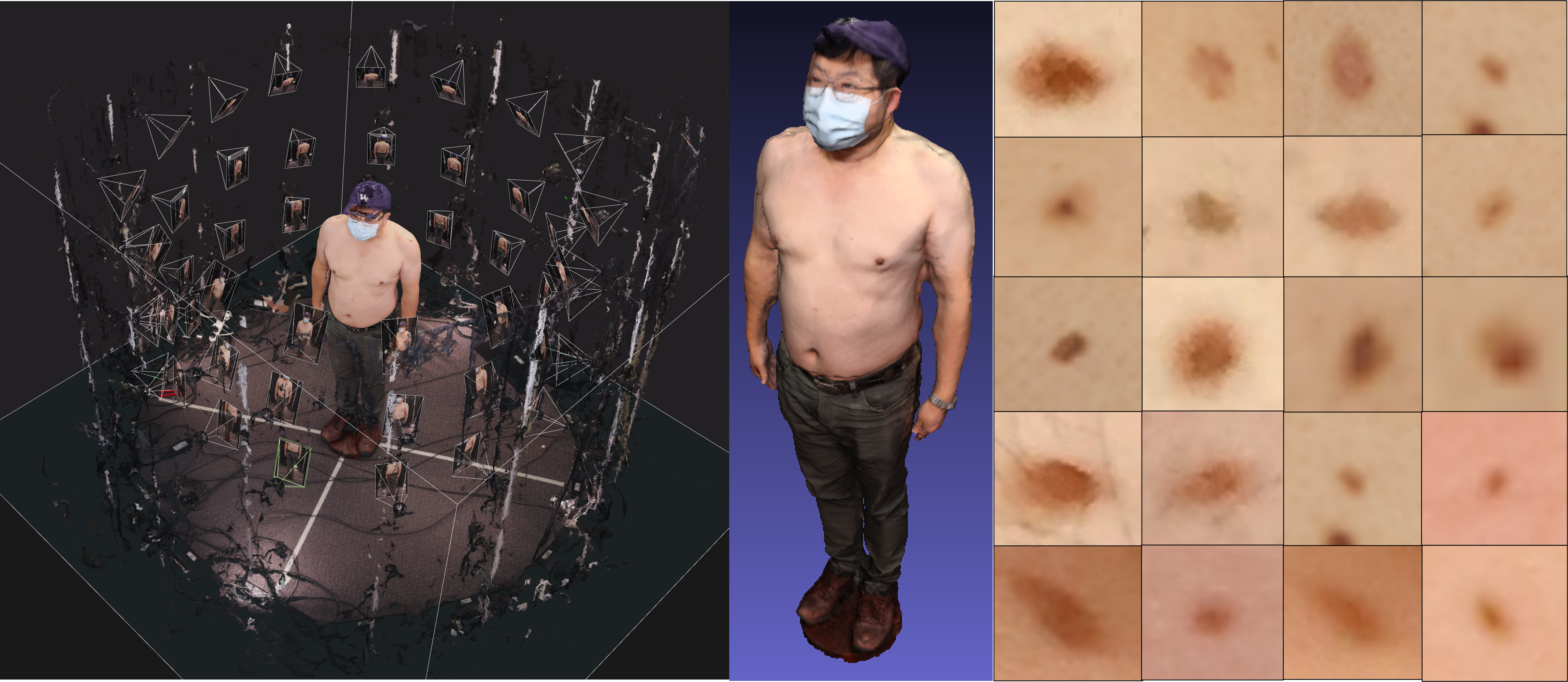}
\end{center}
\vspace{-12pt}
   \caption{3D Whole body imaging captures the entire skin surface to map, document and monitor pigmented lesions. Visualisation of the sparse point cloud (left), 3D texture model (middle), and skin lesions detected (right). All lesions detected on 2D images are mapped to the 3D reconstructed model, improving the ability to track lesions over time. Cameras are shown as 3D wireframe pyramids.
   }
\label{fig:3D_collection}
\vspace{-3pt}
\end{figure}

Although there exist numerous recent articles that demonstrate the benefits of 3D whole body photography~\cite{grochulska2021additive,primiero2019evaluation,navarrete2020total,betz2021reproducible}, 3D human representation is obtained with expensive commercial systems such as the VECTRA{\tiny{\textregistered}}WB360 system (Canfield Scientific Inc)~\cite{canfield}, which simultaneously captures 92 images to reconstruct a 3D avatar~\cite{rayner2018clinical}.
In contrast, we propose a modular mobile health pod equipped with AI based smart tools that can be repurposed from the current clinician use case, to help researchers collect and curate data automatically from 3D human models as illustrated in Fig.~\ref{fig:3D_collection}. 
Existing rotatory solutions for screening (\textit{i.e.} a system that consists of cameras installed on a rotary beam or platform) that have been used previously for skin lesion detection~\cite{strzelecki2021skin} or for the generation of 3D human avatars~\cite{saint20183dbodytex} (\textit{e.g.} Artec Shapify Booth~\cite{Shapify}) are not suitable for clinical purposes because the moving parts of the system are a safety hazard for patients.
It is noteworthy that our system was developed using open-source libraries and software packages for camera control, 3D human body reconstruction, lesion detection, and user interface development, to facilitate greater reproducibility.

With our camera rig, the skin image capture time can be reduced from approximately 6 minutes~\cite{zalaudek2008time, hantirah2010estimating} to a few seconds. After raw data processing by our proposed system, a potential further reduction of examining time can be achieved by more user-friendly data representation generated by the automatic 3D reconstruction and skin lesion detection algorithms. 
Using a 3D record of a patient's entire skin surface, clinical experts can monitor changes in the appearance of lesions that could potentially be early signs of melanoma. The experience and outcome of the patient can be significantly improved by the fact that the doctors will have more time to examine the disease progression and provide more effective treatment~\cite{golda2018recommendations}.

In this paper, we introduce our 3D whole body imaging system to detect and monitor skin lesions.
The contributions of our work are summarised as follows:
\begin{enumerate}
\item A modular image acquisition system has been designed and constructed, enabling researchers to capture images of the whole body of a patient.
\item We have developed a non-contact measurement system that enables efficient documentation, curation and data management of skin lesions compared to traditional manual processes, reducing the time and effort of the clinical expert and improving the experience of the patients.
\item A custom pipeline for 3D reconstruction based on open source tools is proposed to generate high-quality physical-scale 3D body avatars.
\item A processing pipeline that can be used to estimate depth and 3D projection, as well as provide the 3D localisation of skin lesions in whole body images, with a good balance between inference time and accuracy, has been developed.
\item A custom user interface is created that allows users to collaborate and interact with the collected data, the 3D body avatar and detected skin lesions.
\end{enumerate}

\section{Materials and methods} 

\begin{figure*}[!t]
\centering
\includegraphics[width=1\linewidth]{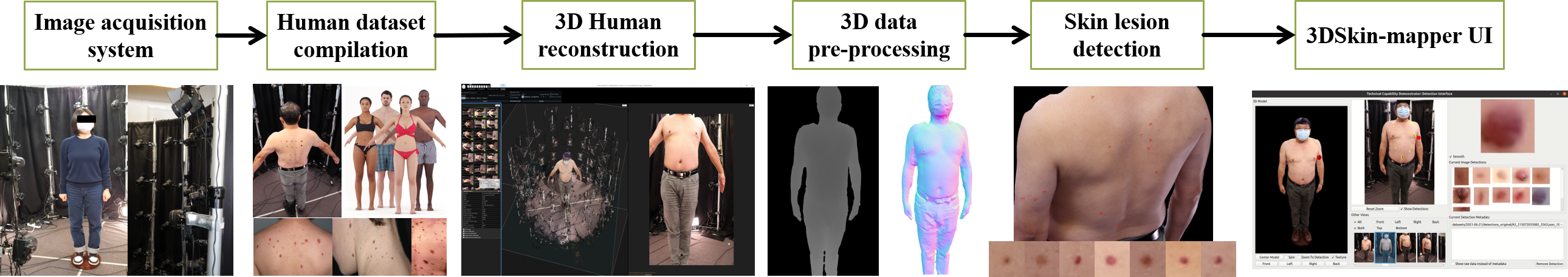}
\vspace{-9pt}
   \caption{Overview of the 3DSkin-mapper workflow.
  \textbf{1.} Design and construction of the hardware and software for the data collection (60 cameras).
  \textbf{2.} Collection of diverse human data for the 3D reconstruction, 3D fine-tuning and skin lesion detection.
  \textbf{3.} Reconstruction of the human body from different camera poses and sparse point cloud estimation.
  \textbf{4.} Estimation of the depth, human mask and 2D to 3D projection.
  \textbf{5.} A deep convolutional neural network that detects lesions from given wide-field images.
  \textbf{6.} 3DSkin-mapper user interface for visualisation, curation and data management.
   }
\label{fig:pipeline}
\end{figure*}

In this paper, we introduce a system called 3DSkin-mapper which performs detection, monitoring and analysis of skin lesions on the patient's entire body.  The system workflow is illustrated in Fig.~\ref{fig:pipeline}.
First, a camera rig with 60 (or more) high-resolution consumer grade cameras is used to capture 2D images of the entire body of a patient simultaneously.
These images are then passed through a processing pipeline for 3D reconstruction, depth post-processing, and 2D to 3D projection. 
A trained deep learning model localises the lesions within the 2D images, which are then mapped back to the 3D geometry of the human body. 
Such anatomical correspondence of the 3D locations of all lesions can be exploited for the longitudinal study of the lesions. 
All information is imported into a user interface (UI) which brings different camera views into a single user point of view.
A key aspect of developing a 3D data collection and curation tool, such as this, is to create an efficient way to facilitate a clinician to capture, analyse and document each of the lesions on the whole body of a patient.
All hardware components used are commercially available off-the-shelf and all algorithms developed are based on open sources to facilitate its reproducibility.
This enables a cost-effective solution with the total hardware cost of less than \$50K AUD.
Details of each module of the imaging system are described in the following subsections.

\subsection{Image acquisition system: hardware and software}
\label{sec:acquisition}

\begin{figure}[!t]
\begin{center}
\includegraphics[width=0.8\linewidth]{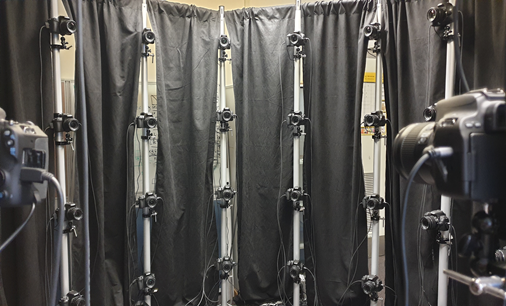}
\end{center}
\vspace{-12pt}
   \caption{Whole body photography structure. Camera cage which consist of 15 poles arranged in a cylindrical configuration, each pole with 4 cameras Canon EOS 200D.
   }
\label{fig:sensors_on_trolley}
\vspace{-4pt}
\end{figure}

The image acquisition system consists of 15 poles approximately 2.1 m tall, arranged in a cylindrical configuration with an approximate diameter of 2.2 m. Four Canon EOS 200D cameras are mounted on each pole at the following distances from the ground: 0.3 m, 0.8 m, 1.3 m, and 1.8 m. Black curtains are placed around the camera rig for privacy, improved lighting as well as blocking out the background of images. 
The cameras are set up to take portrait images with a resolution of $4,000 \times 6,000$ pixels and with the following settings: aperture (F-Stop) = 10, exposure time = 1/15; sensitivity to light (ISO) = 800; and focal length = 18 mm.
During the image acquisition, the patient stands on a motionless wooden platform at the centre of the system, which is used as a reference dimension. This stand is also used to crop and scale the mesh of a patient to the right physical height. The modular acquisition system is shown in Fig.~\ref{fig:sensors_on_trolley}.

\begin{figure}[!t]
\begin{center}
\includegraphics[width=0.8\linewidth]{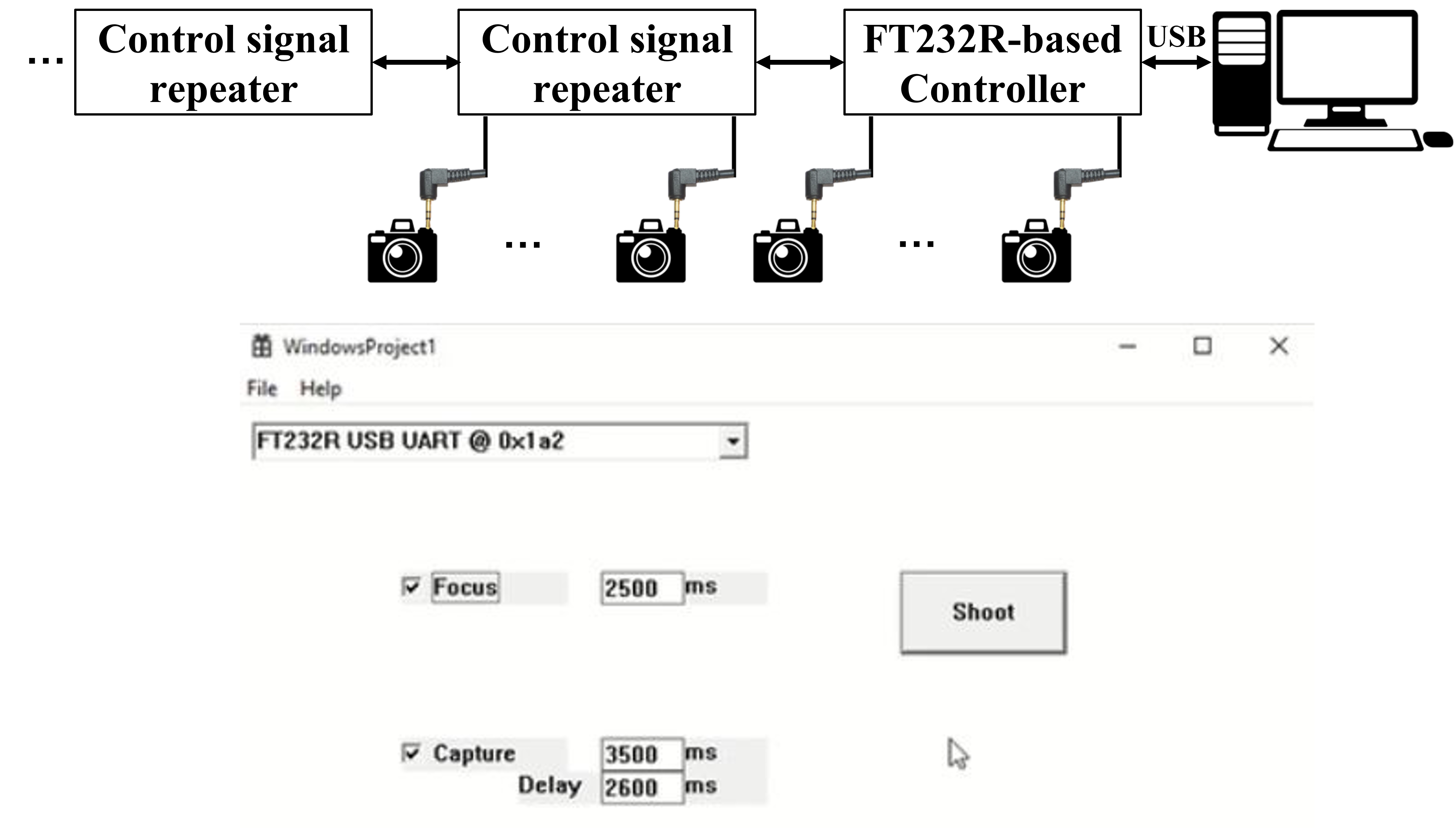}
\end{center}
\vspace{-12pt}
   \caption{Top: Block diagram of the camera triggering circuit. 
            Bottom: Screenshot of the camera control application running on Windows OS.}
\label{fig:trigger_conn}
\vspace{-4pt}
\end{figure}

The Canon EOS 200D cameras used in the system have a 2.5 mm audio jack port for remote triggering. The jack itself has three terminals, which are ground, shutter release, and autofocus, respectively. Connecting either the shutter release or autofocus terminal to the ground terminal will activate the respective function.
Synchronous image capturing is achieved by sending trigger signals to all cameras at the same time. A control board based on FTDI FT232R microcontroller is used for this purpose. The board has 16 3.5 mm audio jack ports for camera connections, and more cameras can be added through purpose-built repeaters. The autofocus signal and shutter release signal are connected to CBUS bit~2 and CBUS bit~3 respectively of the FT232R microcontroller. The signals are set/reset by a Windows application that communicates with the FT232R chip through a USB connection. Fig.~\ref{fig:trigger_conn} (Top) shows the block diagram of the triggering circuit.

\begin{table}[!t]
\caption{Implementation costs of the proposed acquisition system.}
\centering
\resizebox{0.8\textwidth}{!}{%
\label{table:costs}
\begin{tabular}{l c c c} 
\toprule
Component                & Quantity  &  Item Cost  & Line Total \\
\midrule
Canon EOS 200D cameras          & 60        &  700        & 42,000   \\
Pole 2.1m + base plate          & 15        &  80         & 1,200    \\
Black curtains                  & 15        &  20         & 300      \\
Additional lighting             & 4         &  75         & 300     \\
Wooden platform                 & 1         &  30         & 30      \\
Miscelaneous cables $\dagger$   & 1         &  1,200      & 1,200   \\
Control board $\star$           & 1         &  900        & 900     \\
digiCamControl                  & -         &  -          & -      \\
\midrule
\textbf{Total}                  &           &             & \textbf{45,930}      \\
\bottomrule
\multicolumn{4}{p{360pt}}
{
$\dagger$ Cables for remote triggering which includes audio jack port and hubs. \newline
$\star$ Circuit boards and FT232R microcontroller. \newline
Cost in Australian dolars. GST and delivery cost not included.
}
\end{tabular}}
\vspace{-6pt}
\end{table}

The autofocus signal is set first, followed by the shutter release signal. $T_{AF}$, $T_{SR}$, and $D_{SR}$ represent the duration of the autofocus signal, the duration of the shutter release signal, and the delay between asserting the two signals, respectively. 
These parameters can be configured in the Windows application. By default, they are specified to the following values: $T_{AF}$=2,500ms, $T_{SR}$=3,500ms, and $D_{SR}$=2,600ms.
Fig.~\ref{fig:trigger_conn} (Bottom) illustrates the camera control interface. The drop-down menu in the top-left corner of the GUI lists the USB control boards connected to the PC. There should be exactly one device in the list, which is selected automatically when the GUI is launched. The control signals are sent to cameras by clicking the ``Shoot'' button. 
The autofocus signal and shutter-release signal can be enabled or disabled independently by checking/unchecking the ``Focus'' and ``Capture'' boxes correspondingly. This enables different camera settings during image capture. The input boxes next to ``Focus'' and ``Capture'' are used to configure the duration of the autofocus signal and the shutter release signal respectively, while the ``Delay'' input box can be used to configure the delay between asserting the two signals.

The open-source software digiCamControl~\cite{digicamcontrol} is used to change camera settings and transfer captured images from individual cameras to a cloud storage. By launching and properly configuring the software before capturing images, it automatically downloads images from all cameras after new images are captured each time. 
All cameras and image files are labelled according to their locations, which allows easy mapping of image files to the corresponding camera locations in the rig.
The images are captured in less than a second and are then downloaded to the storage.
To facilitate reproducibility of the proposed acquisition system we have provided itemised cost for the components used in the system in Table~\ref{table:costs}.

\subsection{Human dataset compilation}
\label{sec:datasets}

In this section, we provide more details about the data we collected and generated for the training and validation of the components essential for enabling 3D skin lesion localisation and longitudinal tracking. These datasets are summarised in Table~\ref{table:summary_datasets} and are also described in detail in the subsequent subsections.

\begin{table}[!t]
\caption{Overview of the datasets used in the evaluation of our proposed 3DSkin-mapper system.}
\centering
\resizebox{0.98\textwidth}{!}{%
\label{table:summary_datasets}
\begin{tabular}{
lcccl 
>{\raggedright\arraybackslash}p{8cm}
}
\toprule
Component              & Data A     & Data B     & Data C        &Total           & Details\\
\midrule
3D human reconstruction & \checkmark & \checkmark &               & 13 Subjects    & Data A (3 subjects, 2 poses each), Data B (10 subjects, 1 pose each, [2 sources]) \\
Skin lesion detection  & \checkmark  & \checkmark & \checkmark    & 24,525 Images  & Data A (180 images$\star$), Data B (1,632 images,[2 sources]), Data C (22,893, [10 sources]) \\
3D Longitudinal tracking  & \checkmark & \checkmark &               & 130 Lesions    & Data A (3 subjects, 3 sessions, 34 lesions), Data B (7 subjects, 2 sessions, 96 lesions, [1 source]) \\
\bottomrule
\multicolumn{6}{p{520pt}}
{
Data A: Data collected with the proposed imaging system (Discussed in Subsection~\ref{sec:data_collected}) \newline
Data B: Data collected from existing repositories of 3D body scans (Discussed in Subsection~\ref{sec:3D models}) \newline
Data C: Data collected from wide-field of view images (proprietary, open-access repositories, and image synthesis) (Discussed in Subsection~\ref{sec:dataset-skin}) \newline
$\star$ These images were not used in the training of our detector only for validation.
}
\end{tabular}}
\vspace{-2pt}
\end{table}

\subsubsection{Whole body images captured from participants}
\label{sec:data_collected}

Using the proposed imaging system described in Subsection~\ref{sec:acquisition}, we collected images from three (3) healthy participants to test the 3DSkin-mapper workflow - 3D human reconstruction, 3D mapping of skin lesions and longitudinal tracking.
These participants were recruited with the purpose of lesion screening instead of a clinical study.

\begin{figure}[t!]
\begin{center}
\includegraphics[width=0.98\linewidth]{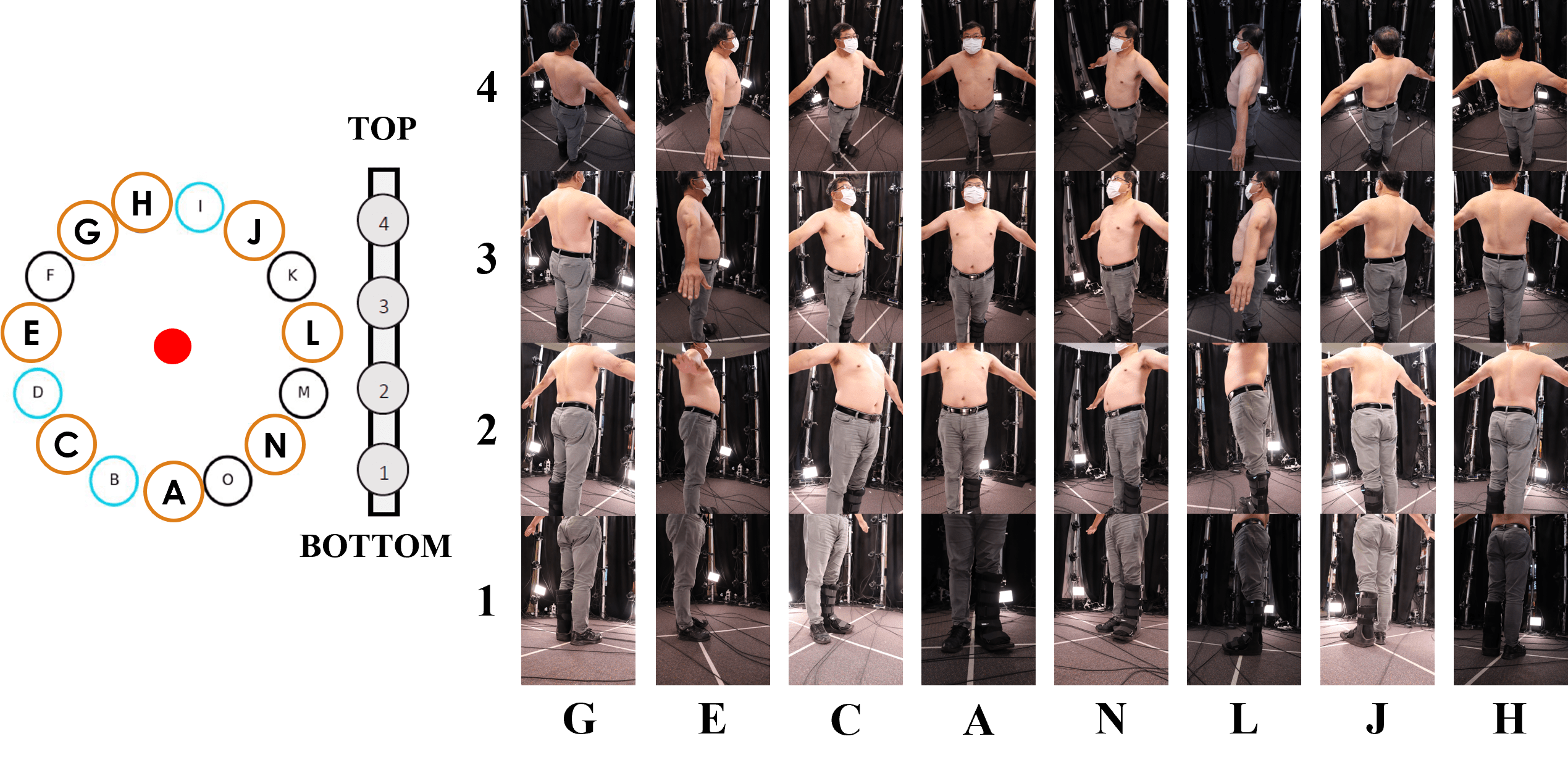}
\end{center}
\vspace{-14pt}
   \caption{Sample images collected from a participant with the proposed whole body imaging system. Selected poles from the imaging system are shown as orange circles.
   }
\label{fig:participant_data}
\vspace{-4pt}
\end{figure}

Participants undergo whole body photography excluding skin on lower body parts (lower limps), feet or scalp, by standing on the motionless wooden platform with a natural standing stance and posture for screening. Each subject was scanned in 2 poses, respectively with arms pointing downwards at an angle (A-pose) and ``arms downward'' without an angle (See Fig.~\ref{fig:pipeline}).
With a single trigger, sixty images are captured by the 60 cameras from different angles simultaneously. These images represent different views of the human body, and are used to reconstruct a 3D human avatar for visualisation, and 3D mapping of skin lesions. Selected views of a participant are illustrated in Fig.~\ref{fig:participant_data}. 

For the longitudinal tracking of skin lesions (\textit{i.e.} to identify changes in size, border, colour, and texture of existing lesions over time) a patient can be scanned at different times to form a set of 3D meshes. 
As a case study, body scans acquired at different times are represented with data collected from these three participants at three sessions: $t1=0$, $t2=15$ days, and $t3=30$ days (\textit{i.e.}~15 days apart from each other).
To evaluate our lesion tracking approach, given two sequential meshes of the same subject $M_{t1}$ and $M_{t2}$ taken at different times, we manually annotate lesions that correspond to each other across the meshes. 
We manually find the location of a lesion in an image from $M_{t2}$ that corresponds to a lesion annotated in an image from $M_{t1}$, and assign both lesions the same unique identifier. We repeat the same process between the images from $M_{t2}$ and $M_{t3}$.
The unique identifier assigned to pairs of matching lesions allows us to determine the ground truth lesion correspondence across meshes. 
A total of 34 unique representative selection of lesions from 3 subjects were manually tracked.
This case study does not examine skin lesions that appear or disappear within sequential sessions as the follow-up sessions were conducted within less than 30 days.
The lesions are initially marked on all 2D images of a single subject to aid in the evaluation and monitoring of the lesion detector, but they are not utilised for training the detector.

\subsubsection{3D body scans dataset from existing repositories}
\label{sec:3D models}

\begin{figure}[t!]
    \centering
    \includegraphics[width=0.8\linewidth]{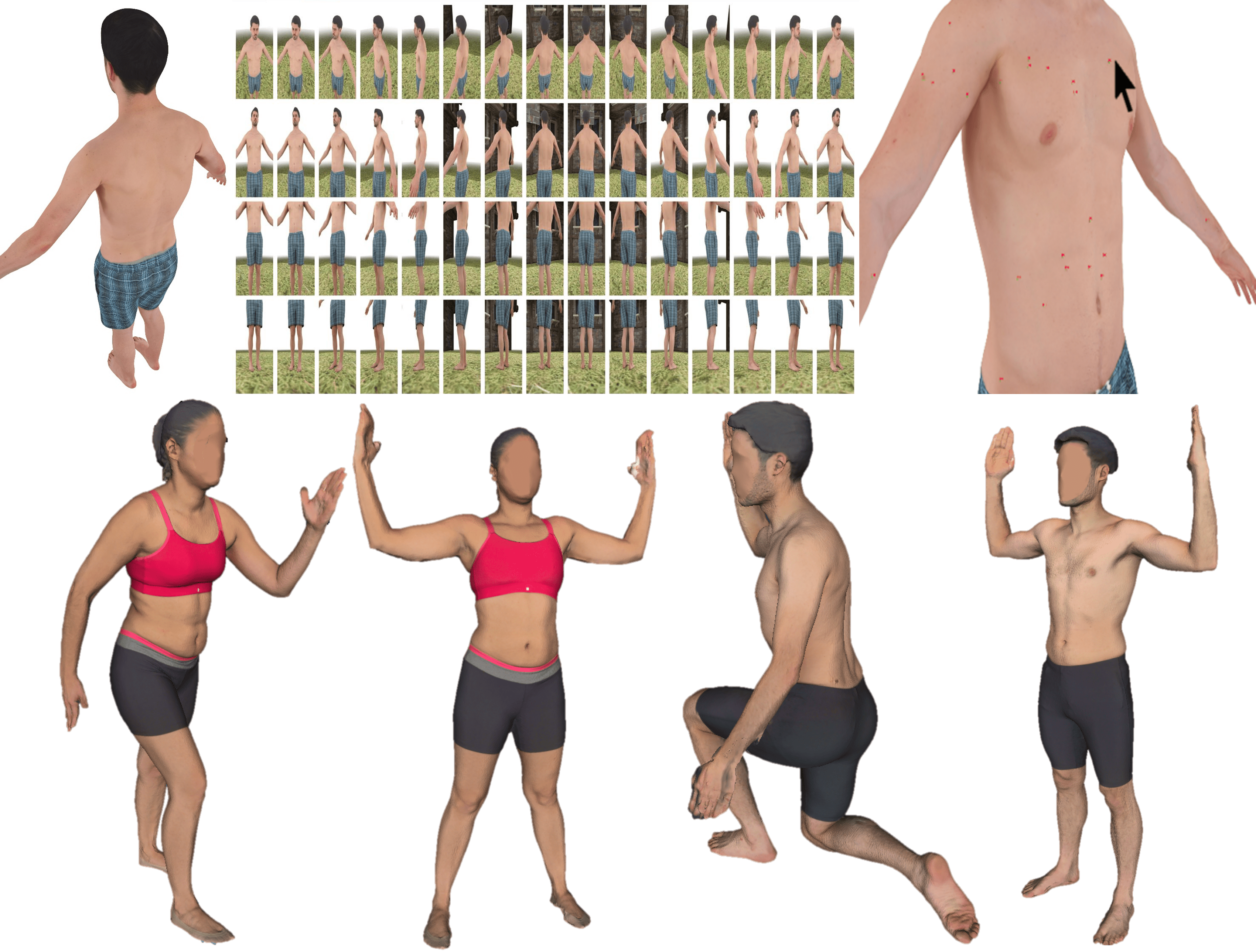}
    \vspace{-7pt}
    \caption{Synthetic 3D human models. 
             Top. Selected human model from Renderpeople~\cite{Renderpeople_bundle_swimwear}, 60 synthetic images generated with background scene and skin lesion visible on the images extracted.
             Bottom. Selected human models from~\cite{saint20183dbodytex,saint2019bodyfitr} with subjects captured in two different poses.}
    \label{fig:3d_human_models}
\vspace{-8pt}
\end{figure}

Various datasets of 3D body scans are available for research or commercial applications. 3D body scan images with high-quality skin texture information are used to optimise camera poses, and test the processing pipeline. Texture information combined with the 3D geometry allows for advanced analysis and representation.
To create realistic images of human figures, we use 3D models from Renderpeople~\cite{Renderpeople_bundle_swimwear} consisting of 4 meshes (4 subjects), and the Texture 3D Body dataset (3DBodyTexV1)~\cite{saint20183dbodytex,saint2019bodyfitr}, utilising 6 randomly chosen meshes (6 subjects). 
From each of the 3D models, we extract 60 synthetic images (with similar configuration as our acquisition system) using Pyrender library\cite{Pyrender} and Blender software~\cite{Blender} for more realistic rendering with a background scene.
Such human models and synthetic images are shown in Fig.~\ref{fig:3d_human_models}.

The synthetic images that were generated, which contain skin lesions, are also included in the skin image analysis dataset discussed in the following section. Additionally, the 3DBodyTexV1 dataset has two scanned meshes of the same subject (captured at different times and poses) which can be used to evaluate longitudinal lesion tracking. 
We selected 7 subjects from the testing set used to validate the lesion detector on 2D images.
Skin lesions from the given pair of meshes $M_{t1}$ and $M_{t2}$ of the same subject are annotated with a unique identifier as explained in Subsection~\ref{sec:data_collected}.
Manual tracking was done for 96 lesions belonging to 7 different subjects.

\subsubsection{Skin image analysis dataset}
\label{sec:dataset-skin}

We curated an annotated dataset on skin disease, which is based on wide-field of view images (captured with digital cameras and cell phones) collected from a commercial lab in Australia, and from open-access 2D image repositories and 3D body scans. Dermoscopic images acquired through a digital dermatoscope, which have relatively low levels of noise and consistent background illumination are not included in our dataset~\cite{hasan2022skin}.
The dataset contains a wide range of skin lesions including common nevus (moles), melanoma and those typical seen in most dermatological patients, \textit{e.g.}~blue nevus, dysplastic nevus (atypical mole), congenital nevus, nevus spilus, angioma, junction nevus, compount nevus, fibroma molle, etc. A nevus that is undergoing changes is a sign of increased risk for melanoma when monitoring a lesion over time~\cite{abbasi2004early}. All skin lesions are categorised as one single class (``pigmented skin lesion'') as our system is designed to capture, analyse and manage the data rather than clinical management of skin lesions~\cite{abhishek2021predicting}.

\begin{figure}[t!]
\begin{center}
\includegraphics[width=0.8\linewidth]{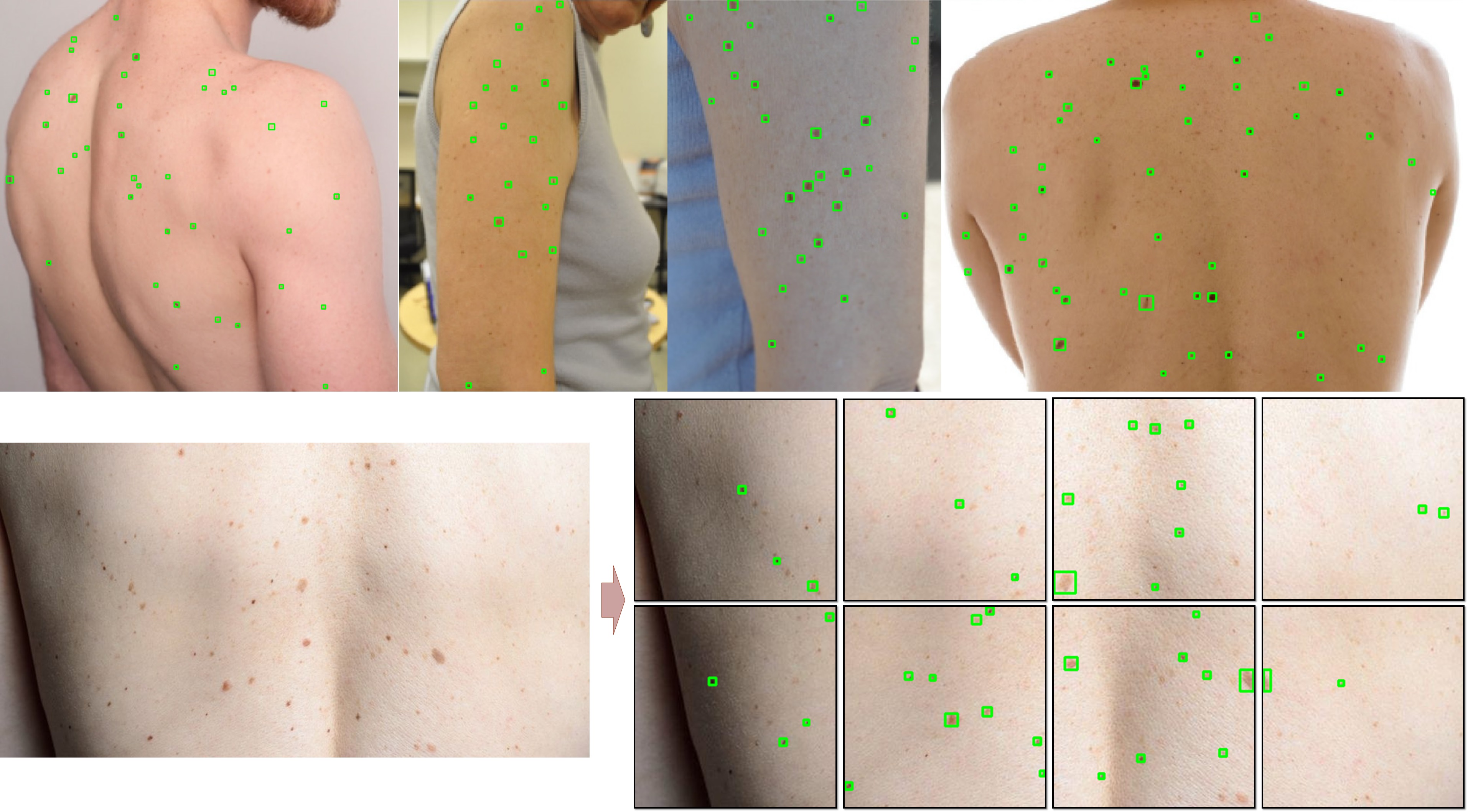}
\end{center}
\vspace{-14pt}
   \caption{
   Top: Selected samples of skin lesions annotated on clinical images.
       Bottom: Each clinical image is split into annotated segments of $608 \times 608$ to train the detector.
   }
\label{fig:dataset-annotation}
\vspace{-4pt}
\end{figure}

\begin{table}[!t]
\caption{Details of the images used to train the skin lesion detector.}
\centering
\resizebox{0.98\textwidth}{!}{%
\label{table:dataset}
\begin{tabular}{l 
c 
>{\raggedright\arraybackslash}p{14cm}
} 
\toprule
\textbf{Dataset} & \textbf{Images} & \textbf{Description}\\
\midrule
ISIC~\cite{rotemberg2021patient}    & 100 
& The 2020 ISIC Grand Challenge contains the largest publicly available collection of quality controlled dermoscopic images of skin lesions. This is the official dataset of the SIIM-ISIC Melanoma Classification Challenge hosted on Kaggle~\cite{international2020siim}.        \\
Dermofit~\cite{ballerini2013color}  & 1,300   
& The Edinburgh Dermofit Library is a collection of focal high quality skin lesion images collected under standardised conditions with internal colour standards. The lesions span across ten different classes including melanomas, seborrhoeic keratosis and basal cell carcinomas.      \\
MED-NODE~\cite{giotis2015med}       & 170   
& MED-NODE contains melanoma and nevus images from the digital image archive of the Department of Dermatology of the University Medical Center Groningen (UMCG) in Netherlands.     \\
SD~\cite{sun2016benchmark,yang2019self} & 3,309 
& SD-datasets (SD-198~\cite{sun2016benchmark} and SD-260~\cite{yang2019self}) contain clinical images collected by digital cameras and mobile phones.\\
PAD-UFES-20~\cite{pacheco2020pad}   & 1,149   
& PAD-UFES-20 dataset consists of samples of six different types of skin lesions. Each sample consists of a clinical image and clinical features including skin lesion location and Fitzpatrick skin type.       \\
Fitzpatrick 17K~\cite{groh2021evaluating}   & 8,289   
& Fitzpatrick 17K dataset contains clinical images with Fitzpatrick skin type labels sourced from two online open-source dermatology atlases: DermaAmin~\cite{dermaamin} and Atlas Dermatologico~\cite{dermatologico}.     \\
MIT dataset~\cite{soenksen2021using}        & 136  
& MIT dataset contains wide-field images extracted from open-access dermatology repositories, web scraping outputs, and de-identified clinical images from the hospital Gregorio Mara\~{n}\'on (Madrid, Spain). We adopted the clinical images provided as supplementary material.     \\
DermNet NZ~\cite{dermnet}           & 428   
& DermNet NZ is supported and contributed by New Zealand Dermatologists on behalf of the New Zealand Dermatological Society Incorporated.      \\
Renderpeople~\cite{Renderpeople_bundle_swimwear}    & 32   
& Renderpeople offers a diverse library of scanned 3D People models. We adopted a bundle of 4 subjects (4 meshes) on swimwear with different skin colour.      \\
3DBodyTexV1~\cite{saint20183dbodytex,saint2019bodyfitr} & 1,600   
& 3DBodyTex consists of 400 high-resolution 3D texture scans of 100 male and 100 female subjects captured in two different poses in finess clothing. Only one mesh per subject was selected. \\
Australian Lab (proprietary)            & 3,535
& We develop an experimental dataset in collaboration with a commercial lab in Australia, which includes de-identified clinical images of skin lesions and ground truth data of the images. \\
GAN-based skin (proprietary)            & 4,487   
&  We adopted GAN-based data augmentation for skin lesion analysis~\cite{bissoto2021gan}. To generate synthetic skin lesion images from our proprietary dataset, we adapted the conditional GAN (cGAN) proposed in~\cite{pix2pix2017} which has been successfully used in many cross-domain learning tasks.   \\
\midrule
\textbf{Total}                      & \textbf{24,525}      \\
\bottomrule
\end{tabular}}
\vspace{-8pt}
\end{table}

\begin{figure}[t!]
\begin{center}
\includegraphics[width=0.8\linewidth]{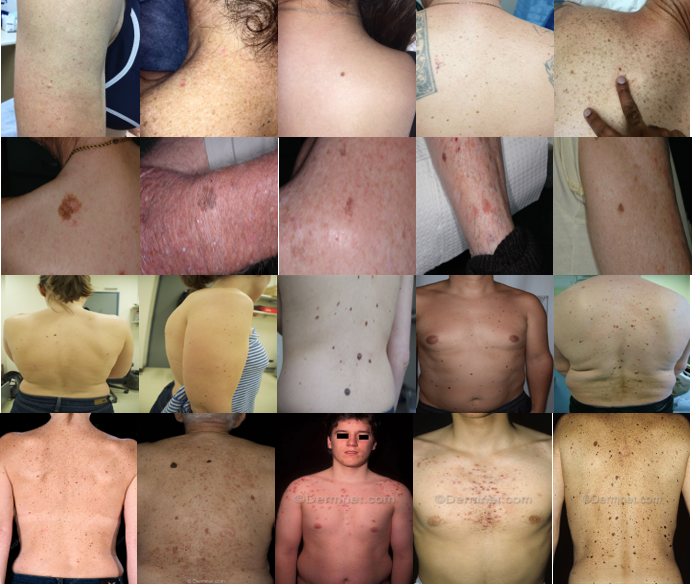}
\end{center}
\vspace{-12pt}
   \caption{
   Selected samples of typical clinical photography included in the skin image analysis dataset. Images from a proprietary source (first row) and publicly available in~\cite{rotemberg2021patient,soenksen2021using,dermnet}.
   }
\label{fig:dataset-collected}
\vspace{-4pt}
\end{figure}

In this work, our objective is to identify the location of a lesion relative to its surrounding skin, so we represent it using a 2D bounding box.
To accelerate the annotation process of 2D bounding boxes for the curation of training data for the lesion detector, a scale-invariant feature transformation (SIFT) is performed first to detect all blob-like regions (group of connected pixels in an image that shares some common property) on the basis of Laplacian of Gaussian (LoG) as suggested in~\cite{soenksen2021using, bogo2014automated}. Skin modelling and analysis of histogram shapes were also considered for lesion localisation.
The preliminary annotations across all images are refined by two human annotators (non-experts) using a computer vision annotation tool (CVAT)~\cite{boris_sekachev_2020_4009388}, and are then reviewed by an expert physician (dermatologist). 
Selected samples of these annotations are depicted in Fig.~\ref{fig:dataset-annotation} (Top).
As the images from different sources have different resolutions, all the annotated images are split into patches (tiles) of $608 \times 608$ pixels which are then used for training. We pad images that have variable dimensions to facilitate model training (See Fig.~\ref{fig:dataset-annotation} Bottom).

Details about the source of the selected images are described in Table~\ref{table:dataset}.
We consolidated a total of 24,525 clinical images annotated. 
From the set of 60 synthetic images rendered from each 3D human model (See Fig.~\ref{fig:3d_human_models} Top), we selected 8 images which correspond to the Top and Bottom views of each position: Front, Left, Right and Back. 
Selected samples of raw wide-field images that contain lesions used to train the skin lesion detector are illustrated in Fig.~\ref{fig:dataset-collected}.

\subsection{3D human reconstruction}
\label{sec:3Dreconstruction}

\subsubsection{Whole body image rendering from 3D human models}
The viewing angles, positions and field of view (FoV), of captured images could significantly affect the quality of the reconstructed 3D human avatar. Therefore, the angles and positions of the cameras were adjusted to capture entire skin surface to construct a complete 3D whole body image with minimal missing parts.
Since we limit the total number of cameras to 60 mounted on 15 poles and setup space inside a lab, the max angles between nearby cameras is 24 degrees and the max height is 2.5m. In addition, the adjustment of the cameras is mostly the height and the tilt angle of the cameras on the poles. In order to expedite the process, we utilised computer graphics rendering to generate synthetic images with a similar configuration and resolution as our camera rig, instead of physically testing different camera positions. Similarly for every camera configuration, a set of 60 synthetic images are rendered from an existing 3D human model from Renderpeople dataset~\cite{Renderpeople_bundle_swimwear}. The images are processed to reconstruct a 3D human model which is visually inspected for quality check. 

There are two conflicting parameters to compromise: i) a wider camera field of view provides a more complete 3D model and more image overlapping between nearby views provides better estimations of camera poses; ii) image resolution for accurate depth estimation for 3D reconstruction.
A camera pose represents the 3D position and orientation of the camera when capturing an image. Incorrect camera poses as well as less view overlapping lead to noisy depth and a low-quality 3D reconstructed model.
In practice, we need to make sure all cameras are detected and used for 3D reconstruction, and minimise the missing part of the output 3D model, particularly the head, shoulder, hands and feet.

Blender~\cite{Blender}, a popular 3D computer graphics open-source software with a Python API, was initially used to render the 3D human model in a controlled environment. However, the rendering time was too long and repeated texture of the environment also caused errors in camera pose estimation during 3D reconstruction.
Thus, we adopt Pyrender~\cite{Pyrender}, a light-weight Python package, to render the 3D model without a background scene. This can speed up the rendering while producing good quality rendered images of the same 24MP resolution as the real captured images for testing the processing pipeline. Another indication of the rendered images with good quality is that the structure-from-motion step in the processing pipeline can detect all the camera poses of the images near expected positions.

\subsubsection{Image acquisition system evaluation}
%
%
In addition to the camera positions, angles and FoV, camera settings including ISO (that is, the gain of the camera sensor), shutter speed/exposure time, aperture and focus also affect the quality of the images. 
Higher gain leads to a brighter but also noisier image. Exposure time represents the amount of light captured by the camera sensor. Longer exposure time leads to a brighter image, but subject to motion blurs. Too high ISO or too long exposure time also leads to image saturation, where image details are lost in very bright image regions. As a result, a balance between ISO and exposure time is determined to obtain the best image quality by visual inspection and checking the image histogram using GIMP software for a given scene and lighting conditions. As the depth of focus is finite and therefore only part of the object is in focus, the focus of the camera lens needs to be adjusted to maximise the image sharpness of the body part to be reconstructed. Bright, minimal saturation, low noise, and sharp images can improve the geometric accuracy of 3D reconstruction.

\begin{figure}[t!]
    \centering
    \includegraphics[width=0.7\linewidth]{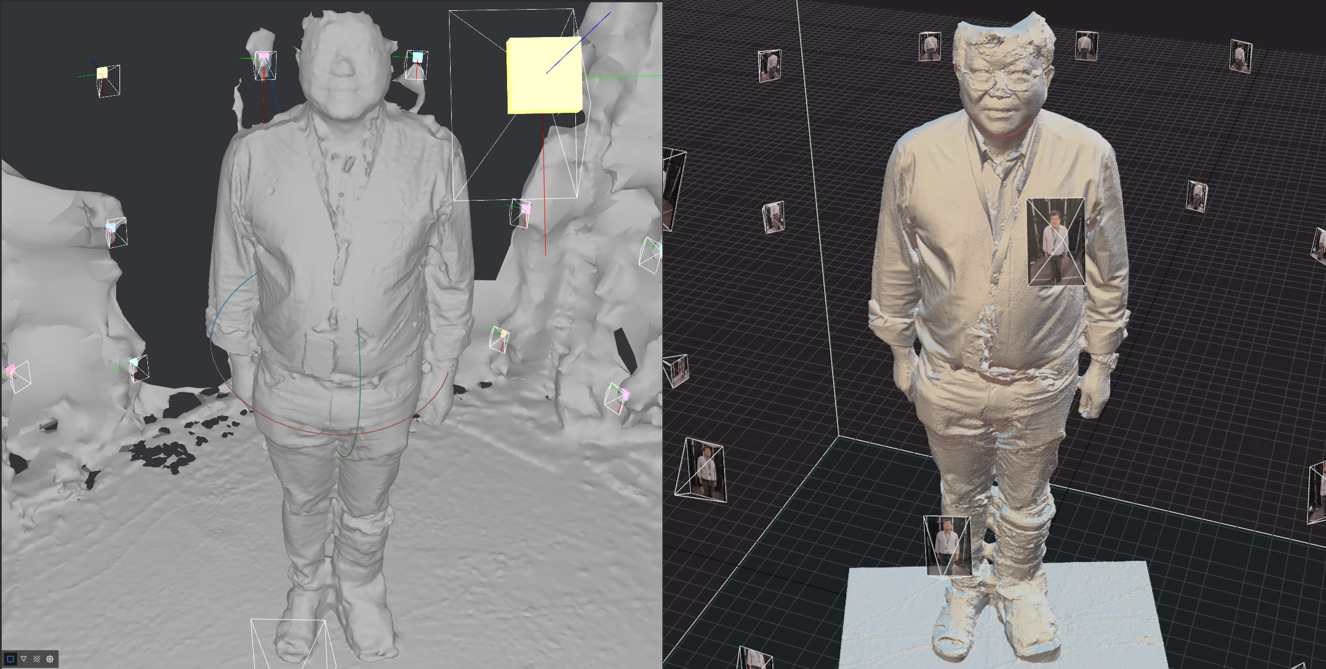}
    \caption{Meshes reconstructed by Meshroom (left) and Reality Capture (right).}
    \label{fig:meshroom_vs_capturereality}
    \vspace{-6pt}
\end{figure}

By optimising these camera settings, we gradually improve the quality of the 3D reconstructed models from real 2D images. Multi-view stereo reconstruction (MVS) successfully detects the camera poses of all 60 images to create a 3D human model as shown in Fig.~\ref{fig:meshroom_vs_capturereality} using two different software solutions. Although the head of the person is incomplete in regions where no lesions are detectable (hair / scalp), the reconstructed 3D mesh is good enough for lesion mapping and analysis.

\subsubsection{3D reconstruction approach}
3D avatars of participants are generated from 3D whole body photography for lesion identification.
For this purpose, two 3D reconstruction software packages were evaluated:
i) Meshroom \cite{Meshroom, griwodz2021alicevision} is a powerful open-source software package that provides very flexible environment for customisation running on Linux and Windows environments; 
ii) Reality Capture \cite{RealityCapture} is a well-known commercial software which provides simple and high-resolution 3D reconstruction running on Windows environment only. The 3D reconstructions were performed on a PC with an i9-9900K CPU, 32 GB RAM, and an RTX 2080 Ti GPU, although lower configuration is also possible as long as a CUDA supported GPU is available.
Fig.~\ref{fig:meshroom_vs_capturereality} illustrates the 3D meshes reconstructed using Meshroom and Reality Capture, respectively.

Both software packages produce decent-quality meshes, however missing a portion of the top of the head. Reality Capture clearly produces a mesh with higher resolution with slightly more completeness. However, Reality Capture was not able to detect the camera poses of four images and excluded them, while Meshroom could detect camera poses for all 60 images. Due to the flexibility and open-source nature of Meshroom, we adopted this tool in the 3DSkin-mapper workflow. For this purpose, we customised a 3D reconstruction pipeline for the implementation of ground plane detection, automatic cropping, scaling and alignment as follows:

\begin{figure}[t!]
    \centering
    \includegraphics[width=1\linewidth]{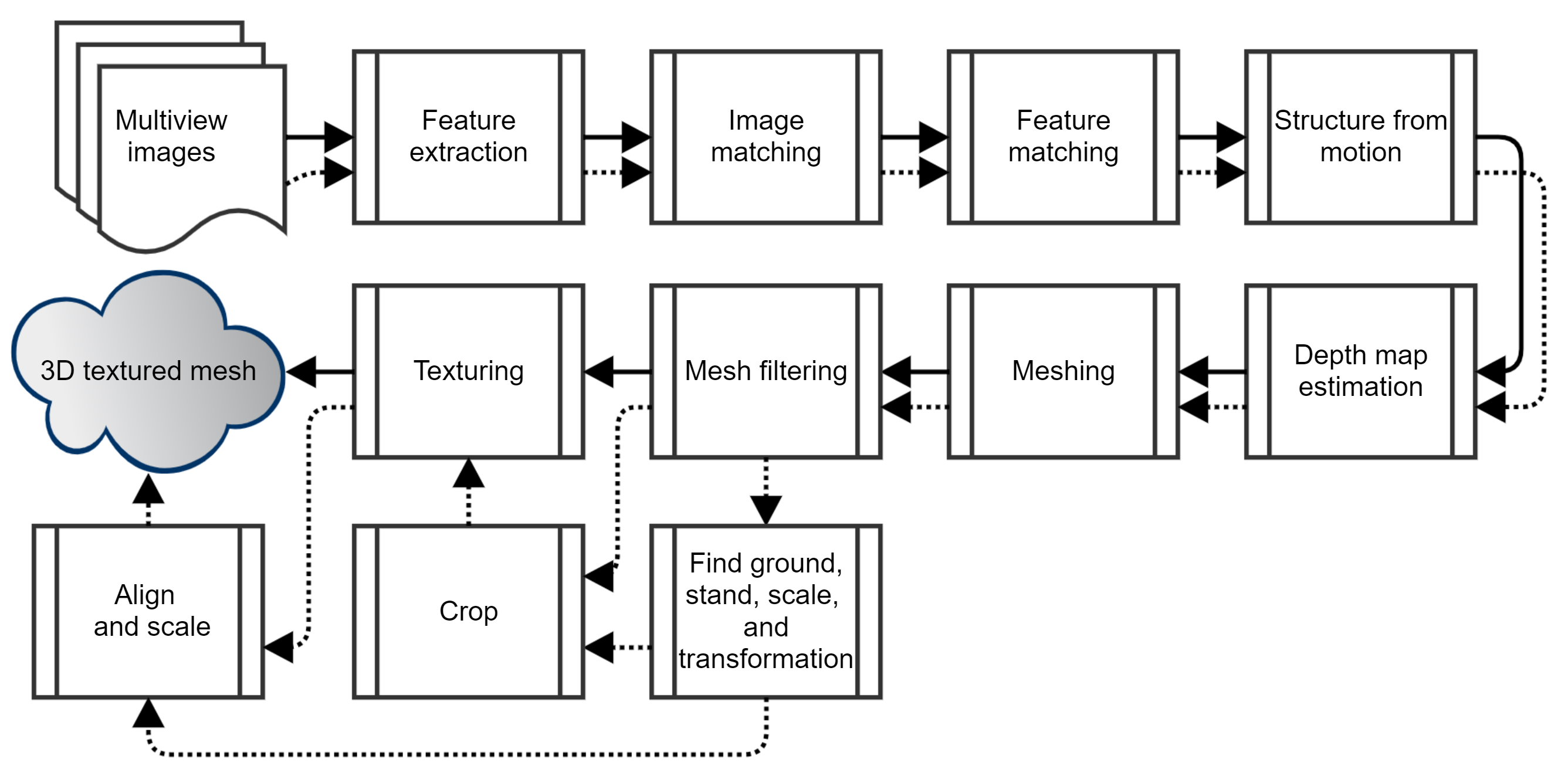}
    \caption{Meshroom's standard 3D reconstruction pipeline with solid arrows. The customised reconstruction pipeline is shown by the dotted arrows to align, crop and scale the 3D mesh.}
    \label{fig:pipelines}
    \vspace{-4pt}
\end{figure}

\begin{figure}[t!]
    \centering
    \includegraphics[width=0.9\linewidth]{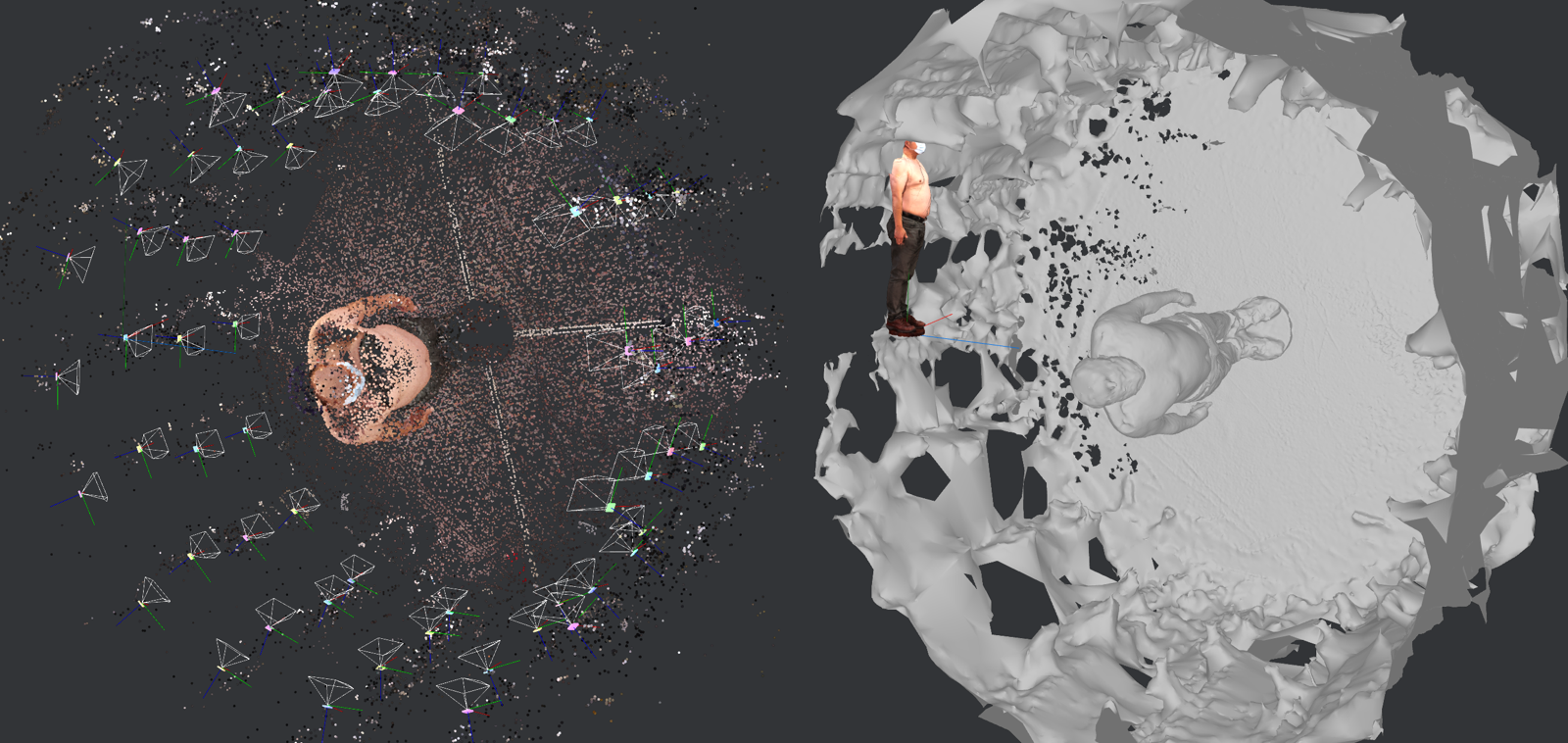}    
    \caption{Camera poses and a sparse point cloud estimated (left), and reconstructed and filtered mesh (right) of the whole scene. The final aligned, cropped, scaled and textured mesh of the person is shown as a small upright-coloured 3D model near the middle. The final 3D model is overlaid together with the filtered mesh to show the differences in size, scale and orientation.}
    \label{fig:overlay}
    \vspace{-6pt}
\end{figure}

\begin{enumerate}
    \vspace{-6pt}
    \item \textit{Standard reconstruction pipeline}. Such a pipeline is illustrated in Fig.~\ref{fig:pipelines} following the solid arrows. The main steps are structure-from-motion, depth map estimation, meshing and texturing. 

    \item \textit{Custom reconstruction pipeline}. The customised 3D reconstruction adds new functionalities including finding the ground (by plane fitting to the reconstructed point cloud), the stand and the scale (by circle fitting) and the transformation of the reconstructed body (by setting the centre and normal vector of the stand to be the world centre and vertical axis respectively).
    Fig.~\ref{fig:overlay} shows the output of structure-from-motion (left), and the filtered non-textured mesh (right) produced by the standard reconstruction. Notice that the non-textured mesh contains the body standing on the floor and the curtain around, and these are removed in the textured mesh of the person.
    The 3D origin is on the ground between the feet, the ground normal represents the y-axis, and the height of the body matches the physical height of the person. The stand provides the scaling factor from the known diameter, and the origin from the fitted circle centre. Such a proposed pipeline is shown as dotted arrows in Fig.~\ref{fig:pipelines}.
    Qualitative results of the 3D whole body skin surface scans and the 3D model with the customised pipeline are depicted in Fig.~\ref{fig:3D_collection}.
\end{enumerate}

\subsection{3D data pre-processing and estimation}
\label{sec:preprocessor}

\begin{figure}[b!]
    \centering
    \includegraphics[width=0.95\linewidth]{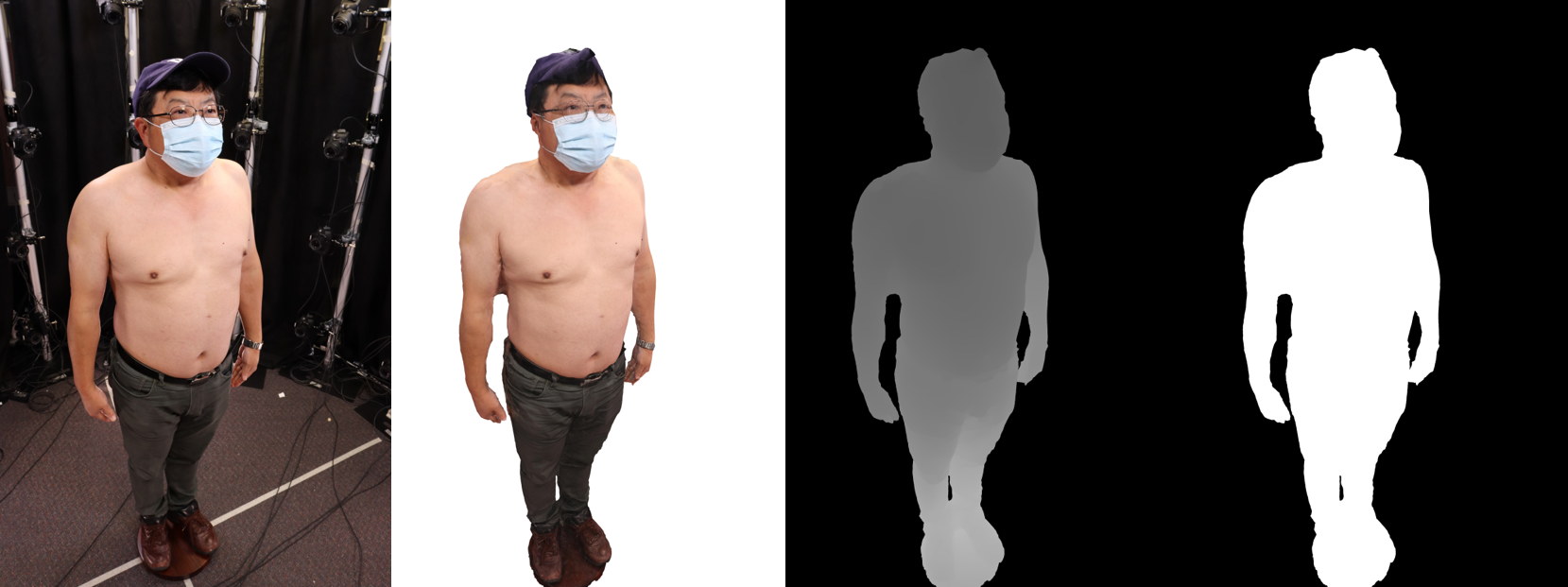}    
    \caption{The input view of preprocessor (left), rendered 3D mesh (middle left), depth image (middle right) and subject mask (right).}
    \label{fig:preprocessor}
    \vspace{-4pt}
\end{figure}

Given the reconstructed 3D model and camera intrinsic and extrinsic parameters, a \textit{preprocessor} renders the 3D models to get the corresponding depth image to each captured image, and a mask indicating which pixels are associated with the subject for each camera view (see Fig.~\ref{fig:preprocessor}). 2D to 3D projection can then be obtained from the depth image and camera parameters. 
These outputs are generated using a process based on 3D rendering for each camera as follows:

\begin{enumerate}
    \vspace{-4pt}
    \item The model view projection (MVP) matrix is calculated from the estimated camera intrinsics / extrinsics.

    \item A scene is constructed consisting of the reconstructed 3D mesh and a single ambient light source. 

    \item The scene is rendered using the Pyrender (OpenGL) library to produce a colour and depth image from the camera's viewpoint.

    \item For each pixel in the image, a 3D ray is calculated from the MVP matrix. The ray is combined with its associated depth value to produce the 3D coordinates associated with each pixel.

    \item Every pixel which has an associated 3D coordinate within the predefined capture region (cylinder with origin in the centre) is marked as part of the subject. 
\end{enumerate}

A small mismatch is present between the input camera view and the rendered outputs, which becomes more pronounced near the edges of the images as shown in Fig.~\ref{fig:preprocessor} (middle left). 
This is caused by the errors from different steps in 3D reconstruction pipeline including lens parameter estimation, depth estimation, 3D meshing, or even missing information from the lack of images and self occlusions. Such errors lead to some uncertainty in the 2D to 3D mapping. Depending on the general distance between nearby skin lesions the uncertainty may not affect that matching of lesions in 3D. As in our limited test cases, such uncertainty does not affect the skin lesion detection in 2D images and their longitudinal tracking in 3D space.

\subsection{Skin lesion detection}

To accurately detect and track lesions on the surface of 3D models over time, we employ a multi-step process. First, lesions are detected on 2D images, and then tracked in 3D using 3D reconstruction techniques. This process allows us to estimate the camera poses, depth, and mask of each input image, which are used to accurately compute the 3D position of each detected lesion on the 3D model. This information is then used for thorough data documentation and longitudinal monitoring.

\subsubsection{Skin lesion detection on 2D images}
\label{sec:lesion-detector}

Human skin lesions are not homogeneous, which include structures such as pathological and benign melanocytic lesions. To identify such lesions, we propose a method based on deep learning as such methods have shown great potential in practical real-world skin lesion examination~\cite{soenksen2021using}. These models can be used to identify the location and determine the dimension (height and width) of a lesion in 2D images collected with the whole body imaging system. The detected skin lesions in the 2D images can be mapped back to the 3D surface of the subject, with the associated 3D coordinates of every pixel estimated with the preprocessor described in Subsection~\ref{sec:preprocessor}.

\begin{figure*}[!t]
\begin{center}
\includegraphics[width=1\linewidth]{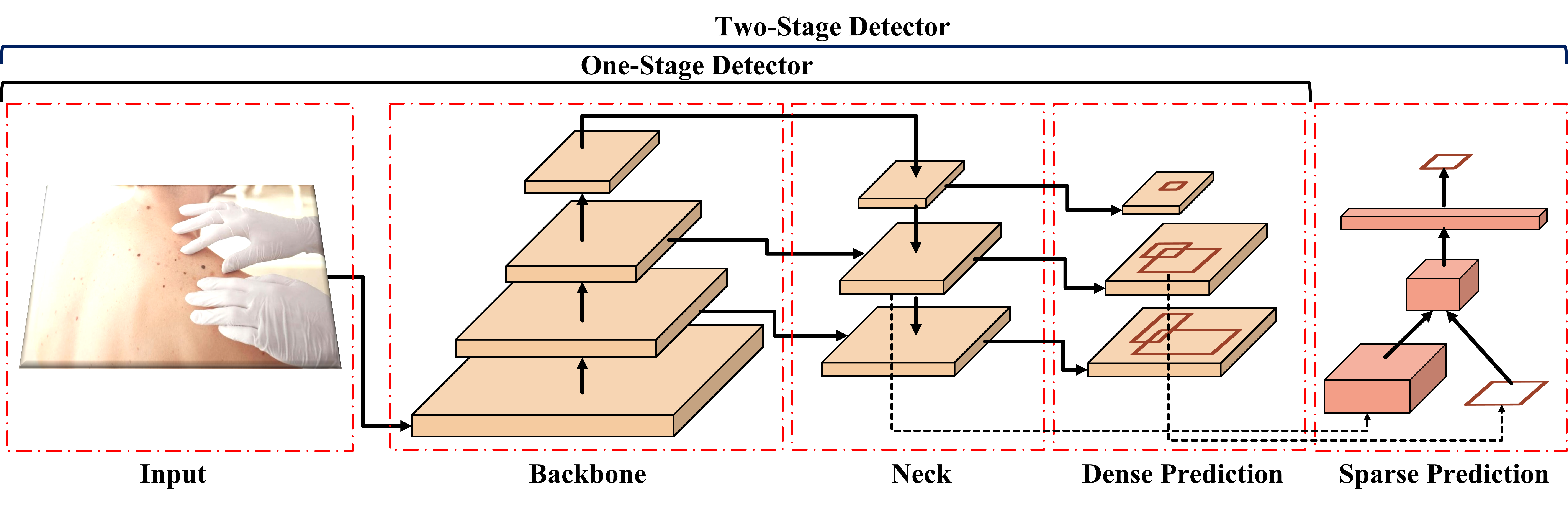}
\end{center}
\vspace{-12pt}
   \caption{Structure of the deep learning based object detectors, highlighting the main components of the Backbone, Neck and Head, and how information flows between these components. Image recreated from~\cite{bochkovskiy2020yolov4}.
   }
\vspace{-4pt}
\label{fig:overview_det}
\end{figure*}

Deep learning based object detectors have achieved exceptional performance in recent years, thanks to advances in deep convolutional neural networks (CNNs). CNN-based models have been widely applied for skin lesion detection and segmentation where pretrained models are employed (on ImageNet\cite{deng2009imagenet}, Microsoft COCO~\cite{lin2014microsoft}) and fine-tuned with the skin lesion datasets~\cite{mirikharaji2022survey}.
The deep learning based object detection models can be classified into two categories: one-stage and two-stage detectors.
One-stage detection frameworks use a single network stage to perform classification and bounding box regression. Two-stage detection frameworks, on the other hand, include a pre-processing step for generating object proposals to identify the object class and to regress an improved bounding box. The structures of both the one-stage and two-stage object detectors are illustrated in Fig.~\ref{fig:overview_det}.

There are many variants of both the one-stage and two-stage models, which are developed by changing the main components of the structures. Common components that may be changed include: the \textit{backbone} used as the feature extractor, the \textit{neck} used to extract different feature maps from different stages of the backbone, and the \textit{head} which is used for the detection of bounding boxes. Such a head can be a dense prediction (one-stage) or a sparse prediction (two-stage)~\cite{liu2020deep,sharma2020comprehensive}.

We opt to evaluate the performance of one-stage detectors as they are better suited and highly viable for real-world applications due to the faster inference than the two-stage detectors. Two-stage detectors such as Faster RCNN used in~\cite{zhao2021detection} for skin lesion detection are popular but require a lot of computational power at inference time.

Among well-known \textit{one-stage object detectors}, the scaling cross-stage partial network (\textit{Scaled-YOLOv4})~\cite{wang2021scaled} from the YOLO family of detectors (You Only Look Once~\cite{redmon2016you}), can achieve competitive accuracy while maintaining a high processing frame rate with an inference speed greater than 30FPS.
Such an accurate and efficient model has several features that make it well-suited for commercial applications. Firstly, it is relatively easy to train and deploy, and it has a relatively small model size, making it more suitable for the deployment on resource-constrained devices. Secondly, it can process multiple scales of input images in a single pass and perform real-time object detection on high-resolution images. Lastly, its implementation with the Darknet deep learning framework available in~\cite{darknet-ab} allows for commercial use.

Scaled-YOLOv4 makes an incremental improvement to YOLOv4~\cite{bochkovskiy2020yolov4} with a network scaling approach that modifies not only the network depth, width, and resolution; but also the structure of the network. 
The model is composed of an optimised CSPDarknet53 as its backbone, which employs a Darknet-53~\cite{redmon2018yolov3} and a Cross Stage Partial Network (CSPNet) strategy~\cite{wang2019cspnet} to partition the feature map of the base layer into two parts and then merge them through a cross-stage hierarchy. An additional spatial pyramid pooling (SPP)~\cite{he2015spatial} module and a path aggregation network (PANet)~\cite{liu2018path} with CSP are used as the neck. Finally, a YOLOv3~\cite{redmon2018yolov3} is used as the head of the model.

\paragraph{Model training and experimental setup} \hfill

A Scaled-YOLOv4 model pretrained on the MS COCO dataset is fine-tuned with the Darknet framework on our skin lesion dataset, as described in Subsection~\ref{sec:dataset-skin} (clinical images spanning various skin-types and body locations. The 24,525 images are randomly split into training (14,715 images), validation (3,924 images), and testing (5,886 images) sets. 
To prevent bias against certain groups of people, images from the PAD-UFES-20 and Fitzpatrick 17K datasets were selected based on their skin type labels for the training, validation, and testing sets. Additionally, male and female subjects from the 3DBodyTex dataset were evenly distributed in the training dataset.
The model is trained for 10,000 iterations, with a batch size of 64, a learning rate of 1$e^{-3}$, a loss function of sum-squared error (SSE), and a bounding box regression loss CIoU~\cite{zheng2020distance}. In Darknet, an iteration refers to the process of adjusting the model's parameters using the gradients calculated from a batch of training data.
The training dataset is enhanced by using various data augmentation techniques such as Mosaic, randomly rotating images between 0 and 15 degrees, randomly modifying the brightness, contrast, saturation, and hue of each image, and randomly flipping the image horizontally.

To enhance the prediction performance of the skin lesion detector, the whole body images are split into patches of $608 \times 608$ pixels with an overlap of 50\%. The detected lesions are then aggregated across all patches and non-max suppression (soft-NMS~\cite{bodla2017soft}) is applied to obtain the final set of lesions detected for the input image.

\subsubsection{2D matching and 3D mapping of skin lesions} 
\label{sec:2D-3D mapping lesion}

The main challenge that our acquisition system addresses is the association of lesions, or matching a lesion to all images in which it is identified. During image acquisition, the scanner takes a series of overlapping images to ensure that each lesion is captured from at least 5-6 different viewing angles of cameras mounted on the poles of the acquisition system. 
In order to avoid repeatedly analysing the same skin lesion across multiple images, we have developed a method that effectively matches detected lesions and establishes the correspondence between each lesion and its respective images. This approach aims to prevent duplication of effort in lesion analysis.

To address the matching problem, we utilise the 3D position of the lesion to more accurately determine the correspondence. Our whole body imaging system captures a set of images and their corresponding camera poses to evaluate the 2D-3D mapping of the detected lesions visible in multiple images. 
We first map the centre point of each predicted 2D bounding box to 3D space as the 3D coordinates associated with each pixel are known (See Subsection~\ref{sec:preprocessor}). As a lesion can appear multiple times in various images, skin lesions with similar 3D positions on the 3D mesh are grouped together using a unique global lesion ID. 
When it comes to 3D points, a distance threshold is used to specify the maximum distance between two points for them to be considered part of the same cluster.
Using this 3D approach to align lesions across images, it has shown better performance than traditional 2D methods that depend on geometric constraints, 2D templates and graph matching~\cite{bogo2014automated}.

\begin{figure*}[!t]
\begin{center}
\includegraphics[width=0.9\linewidth]{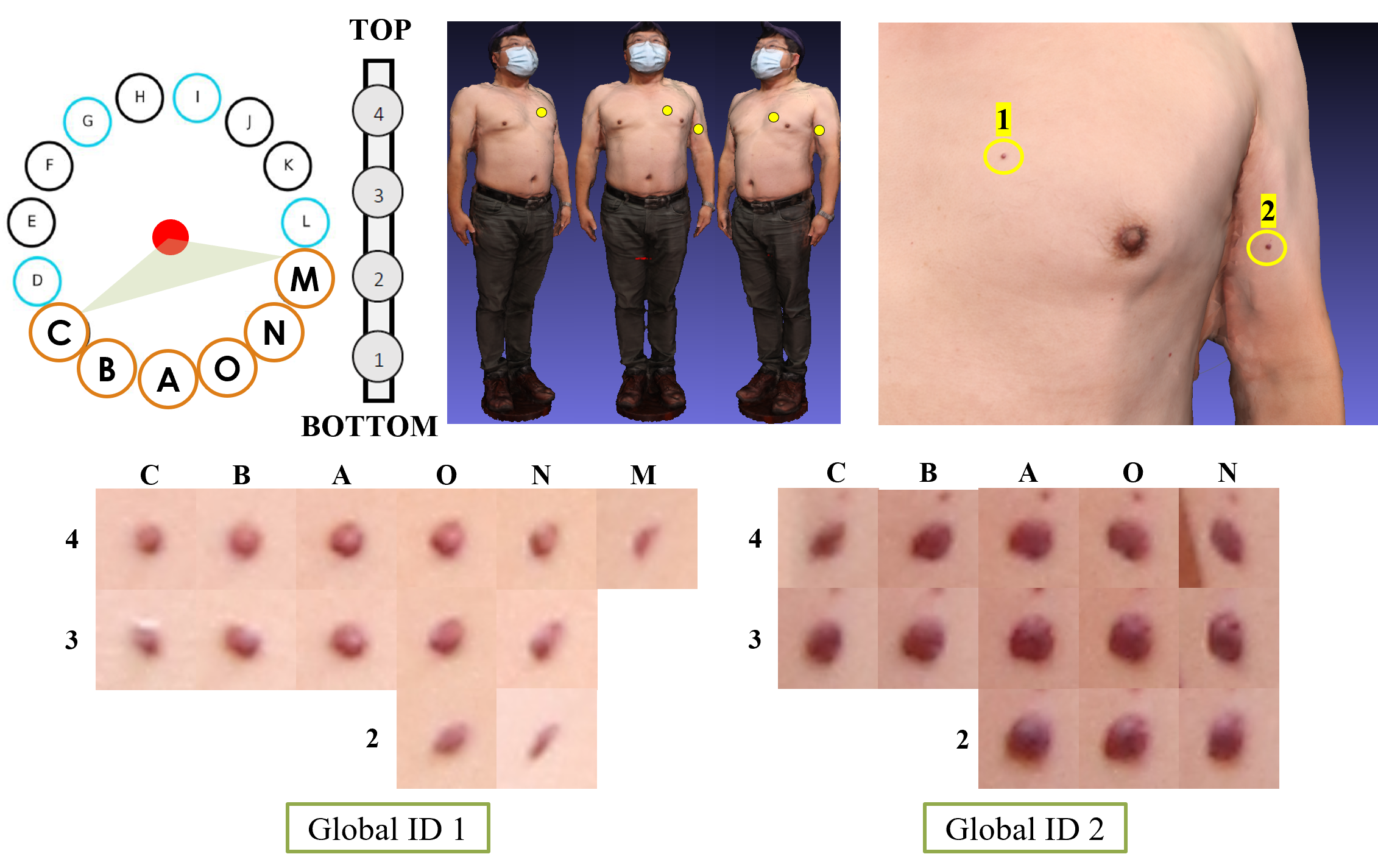}
\end{center}
\vspace{-12pt}
   \caption{Visualisation of two selected skin lesions and their different views from each camera. Lesions visible from poles C, B, A, O, N, and M.
   }
\label{fig:global_ID}
\vspace{-6pt}
\end{figure*}

In order to achieve this objective, we employ an agglomerative clustering technique that is computed using the SciKit-learn library~\cite{scikit-learn}), with a thresholded distance of 0.02 and Euclidean metric which is used to measure the distance between instances in a feature array. This parameter was empirically set at 0.2 as it yields the highest accuracy to define the clusters taking into account the size of the images ($4,000 \times 6,000$ pixels).
Our framework that uses multiple cameras is utilised to eliminate lesions that are not consistently identified by a series of cameras, which can be considered as outliers or possibly incorrect detection.
Agglomerative clustering groups similar points together by repeatedly combining smaller clusters into larger ones. Each point starts at its own cluster, and the closest pair of clusters are joined together until all points are in one cluster or a predetermined stopping point is reached.
The model provides the clusters with each data point belonging to them after running the hierarchical clustering algorithm. Clusters of less than three (3) points are rejected as outliers based on the overlapping between nearby views in the acquisition system. The number of clusters is equivalent to the number of unique lesions, and the 3D coordinate of the global lesion ID is defined as the centroid of the cluster.
Fig.~\ref{fig:global_ID} illustrates the different viewing angles of two selected global lesion IDs and their representations in the 3D space.

\subsubsection{Anatomical correspondence and longitudinal tracking} 
\label{sec:tracking}

When studying two images that depict the same lesion before and after a change, a method is required to confirm that they represent the same lesion during a temporal analysis. To study the progression of lesions over time, we take into consideration the differences in the pose between scans.
The 3D information is used for matching and comparison as in clinical practice the visual appearance of the lesions change over time in different scans, which is a limitation of early methods based on only 2D anatomical landmarks and structured graphical models~\cite{mirzaalian2016skin, korotkov2018improved}.
Comparing to using multiple single images, detected lesions integrated into a full-body 3D map greatly simplify the tracking of changes in lesions.
For this purpose, we first determine anatomical correspondence across 3D human meshes, we then establish the correspondence between the lesions (3D spatial coordinates) on two sequential meshes~\cite{bogo2014automated}.

Learning-based deformation methods can be used for 3D correspondence matching between two 3D meshes~\cite{deprelle2019learning}. This ensures that skin textures captured at different times are precisely aligned, making them directly comparable. This makes it easy to evaluate the changes in lesions over time.
To generate vertex correspondence across scans of the same subject (per pairs), we adopted the semi-supervised scan registration method LoopReg~\cite{bhatnagar2020loopreg} trained on the Faust dataset~\cite{bogo2014faust}, which contains 100 scans of undressed people in challenging poses and noisy scans, and corresponding skinned multi-person linear model (SMPL) registration~\cite{loper2015smpl} as ground truth.   
Such a model does not require instance specific annotations or multiple initialisations in comparison with supervised models such as 3D-CODED~\cite{groueix20183d} adopted by~\cite{zhao2021detection} to register skin meshes. Details of the LoopReg approach are provided in~\cite{bhatnagar2020loopreg}.

With the vertices of two registered meshes of the same subject using LoopReg, obtained at different scan times, we can now proceed with the lesion matching across meshes. 
The anatomical correspondences between two temporally adjacent meshes are used to register the 3D coordinates of the lesions.
We map the 3D position of each lesion, which is represented by a unique global lesion ID as discussed previously, to the closest vertex on each of the corresponding 3D meshes of the same subject.
Similar to previous works~\cite{mirzaalian2016skin}, let $l_t$ and $l_{t+1}$ be lesions from two scans ($M_{t1}$ and $M_{t2}$), the matching function can be formalised as $M : (l_t, l_{t+1}) \rightarrow \{0,1\}$ which provides $1$ when $l_t$ and $l_{t+1}$ correspond to the same lesion, and $0$ otherwise.

We then perform lesion tracking by determining the 3D geodesic distance between the vertices of the registered meshes to correct for any inaccuracies in the mesh registration process. We assume that the geodesic distance remains relatively consistent across all poses which allows us to locate the corresponding lesion in the closest anatomical location of the target mesh. 
This operation is accomplished through the Dijkstra's algorithm. The algorithm finds the shortest path between two vertices on a mesh by treating it as a graph with vertices as nodes and edges as paths. The algorithm begins at the starting vertex, explores its neighbouring vertices, and records the distance for the starting vertex to each one. It then selects the unvisited vertex closest to the starting vertex and continues the process until the target vertex is reached~\cite{crane2020survey}.
Dijkstra's algorithm approximates the actual geodesic distance with the length of the shortest piecewise linear path on mesh vertices.
Given $\upsilon_{t}$ and $\upsilon_{t+1}$, two vertices of both adjacent lesions $l_t$ and $l_{t+1}$, $P(\upsilon_{t},\upsilon_{t+1})$ a path connecting the two vertices, and $L(P(\upsilon_{t},\upsilon_{t+1}))$ the path length, the geodesic distance $g3d$ between two vertices is approximated by the following expression: $g3d(\upsilon_{t},\upsilon_{t+1})= \nicefrac{min}{P}L(P(\upsilon_{t},\upsilon_{t+1}))$.
Such formulation assumes that the assumption being made is that the number of skin lesions observed in $M_{t1}$ and $M_{t2}$ are equal, which is consistent with the specific case being studied where the lesions do not change significantly in a short period of time between follow-up sessions.

\subsection{3DSkin-mapper UI}
\label{sec:ui}

\begin{figure}[!t]
\begin{center}
\includegraphics[width=0.96\linewidth]{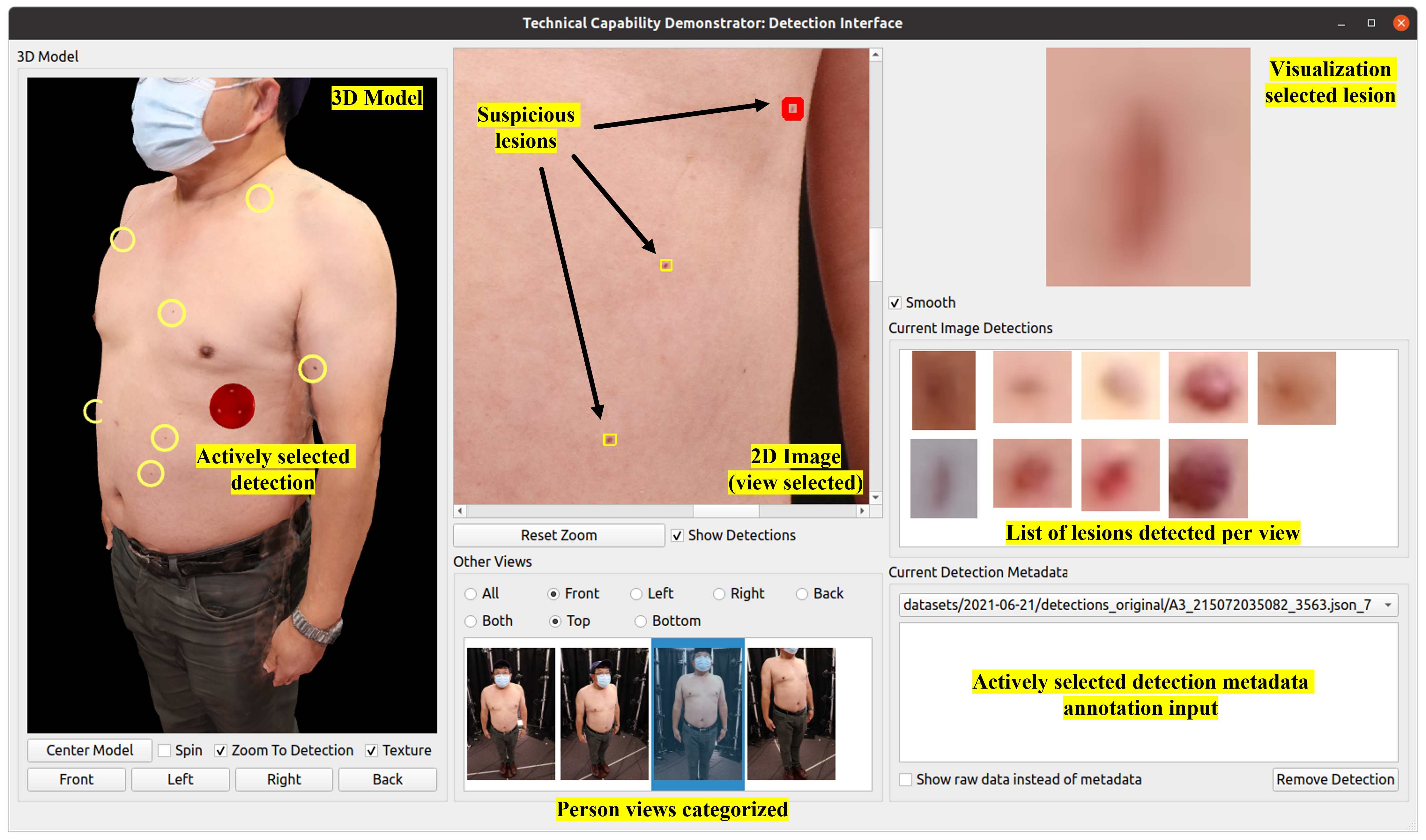}
\end{center}
\vspace{-12pt}
   \caption{User interface for the proposed system. The cropped images on the right that contain each lesion look blurry as they are zoomed-in and at the limit of image resolution.
   }
\label{fig:qualitative_results_interface}
\vspace{-6pt}
\end{figure}

We developed a custom user interface (UI) to visualise our results. As illustrated in Fig.~\ref{fig:qualitative_results_interface}, the interface centres around multiple linked views of the captured and processed images.

A thumbnail grid view of the images that were captured using the imaging system (See Fig.~\ref{fig:participant_data}) appears at the bottom of the central column of the interface. Users can filter the visible thumbnails based on the area and pole position the images were captured from (\textit{e.g.} Front view: Poles C, B, A, O and N). 
The selected image appears in the central top view of the interface and allows the users to zoom and pan around the image for more detailed inspection. The rectangular region of the detected features can be shown as an overlay on the central primary image. An individual lesion can be selected within the bounding box region of the lesion on the central primary image.

The left side of the interface shows the reconstructed textured 3D model (See Subsection~\ref{sec:3Dreconstruction}). The view of the 3D model can be interacted with a mouse, allowing the model to be rotated around the centre of the model (mid-abdomen), as well as manipulating the camera around the model (pan / tilt as well as moving the camera in/out). There are also user interface options to align the 3D camera view to a fixed predefined position pointing towards the model, as well as moving the model back to the centre of the viewport (while maintaining the current camera position). The model appears with textures by default, but the textures can be turned off to allow more detailed inspection of the 3D mesh without the added visual cues from the textures, which mask some underlying 3D model imperfections.

The right panel highlights the active detection (top-right) showing either the raw image pixels of the detection enlarged with smoothing optionally applied to facilitate clinical use cases. All detected lesions from the current image appear in the ``Current Image Detections'' image grid. The ``Current Detection Metadata'' region can be used by the user to annotate the currently selected lesion with free text notes.
Further, we allow for curation of the dataset by annotating misdetections with the ``Remove Detection'' button (the selected lesion is then removed from visualisation on the user interface). Selecting the active detection (either by selecting within the bounding box region of the lesion directly from the central primary image or the ``Current Image Detections box'') updates the selection across all views (highlighting on the 3D model, primary image, enlarged image crop and selected current image detection and metadata). 
When the actively selected detection is changed, the virtual camera viewing the 3D model can optionally be moved to a fixed offset from the 3D point on the mesh surface of the model to point directly at the computed 3D points of the detection (focusing the 3D view on the actively selected detection). This is achieved by offsetting the camera a fixed distance from the computed 3D coordinates of the detection on the model surface along the computed normal vector at this location. 

The UI was developed so that it was agnostic to the underlying model generation, skin lesion detection and geometric transformation computation. This completely decouples the data processing from the user interface presentation, by using directories of flat files as the transport mechanism between these separate systems. This means that the UI is a modular product that could be reused for a different application by simply feeding in the appropriate model, images, and detection metadata.

The UI was developed using QT via PySide2 bindings from Python version 5.15.2. The UI takes as input the 60 images saved from the DSLR cameras, the 3D model (stored in an .obj file with an associated single texture image) and a JSON data file which contains the information on the detections (\textit{e.g.} unique identifier, 3D coordinate location and 3D normal direction). Linkage of the data in the JSON file to the underlying image in which the detection is found can be made via filename correspondence (prefixed the same - different file extensions) or a metadata field which stores the original image in which the lesions are detected.

The UI provides an alternative but effective way to present skin information to doctors. It allows doctors to see skin lesions virtually in their 3D positions, and easily track them over time. 

\section{Results} 

\subsection{3D model reconstruction}

\begin{figure}[!t]
\begin{center}
\includegraphics[width=1\linewidth]{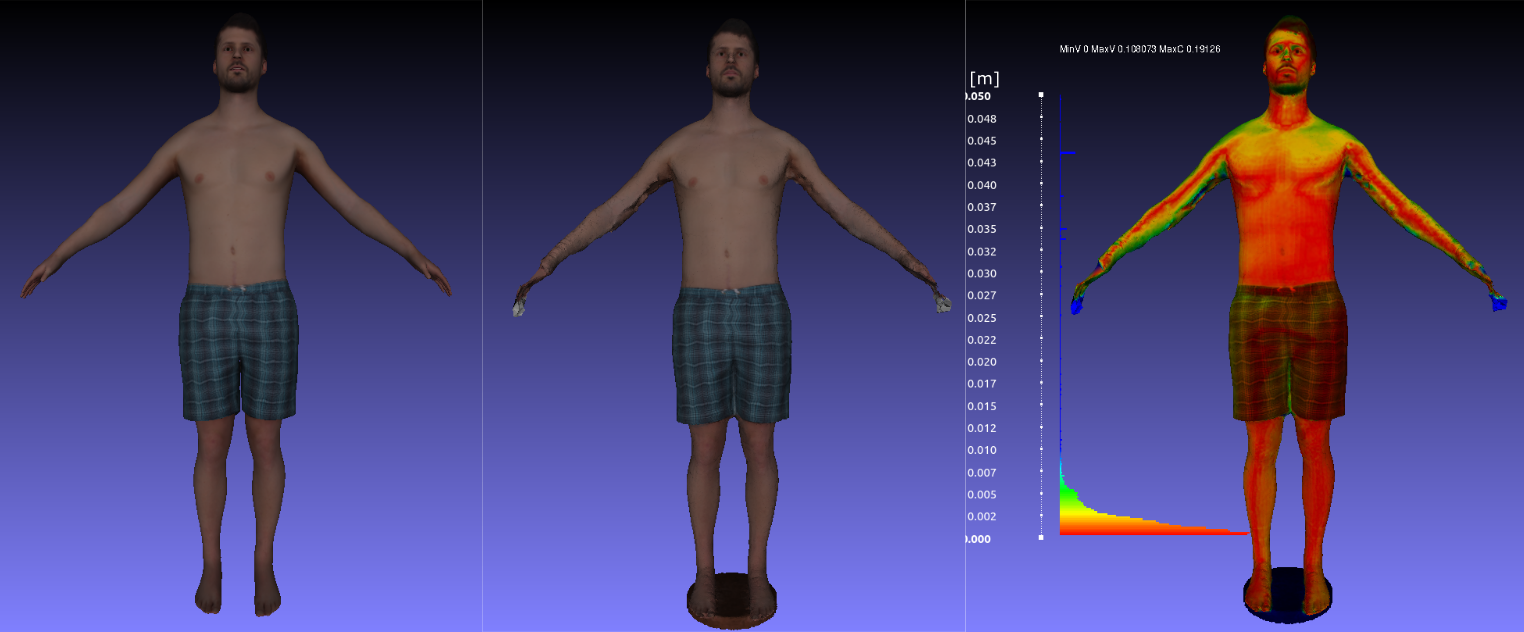}
\end{center}
\vspace{-9pt}
   \caption{Comparison between the ground-truth human model (left) with the reconstructed model (middle). Synthetic images generated from the ground-truth model were used to produce the reconstructed model. The reconstruction error in terms of Hausdorff distance is shown as colour map and histogram (right). The model displays some missing elements in the hand and underarm areas, with an average error of approximately 2 mm. 
   }
\label{fig:comparison_groundtruth_reconstruction}
\end{figure}

Validation of 3D reconstruction accuracy is performed by computing the Hausdorff distance (using MeshLab~\cite{cignoni2008meshlab}) of the reconstructed 3D mesh relative to the ground truth 3D mesh (from Renderpeople and 3DBodyTexV1) which is used to generate synthetic images for 3D reconstruction. 
Hausdorff distance is defined as the largest distance between two meshes. When comparing two meshes, large Hausdorff distances indicate missing features in the reconstructed mesh. Once the Hausdorff distance is computed and normalised by the diagonal distance of the 3D mesh bounding box, it is visualised as a heatmap on the ground truth mesh.

The results and their comparison to a selected ground truth 3D mesh from Renderpeople are shown in Fig.~\ref{fig:comparison_groundtruth_reconstruction}. An average error across the 10 models introduced in Subsection~\ref{sec:3D models} was 2.5 mm. However, errors were concentrated in the areas of limited image coverage such as underarm and upper shoulder. The entire 3D reconstruction process takes approximately 30 minutes or less.

\subsection{Lesion detection and tracking}

\subsubsection{Detecting lesions on 2D images and 3D mapping}

The accuracy of the lesion detection is evaluated by comparing the 2D bounding box predicted by our method (that encapsulates the pigmented skin lesion) to the corresponding manual annotation (ground truth) in the image in which the lesion is detected, as described in the Subsection~\ref{sec:dataset-skin}.

We use the normalised intersection-over-union (IoU) metric to define how well a bounding box overlaps the associated ground truth. We adopt the mAP@IoU[0.5], where the mean average precision (mAP) requires that the bounding boxes overlap with an IoU $\ge$ 0.5. 
We also report recall and precision. Precision represents the proportion of true positives among all positive predictions, while recall represents the proportion of true positives among all actual positive instances. We calculated each metric for all images and report the average over all images.
It is worth noting that, while high precision is preferred, high recall is crucial as the consequences of missing a potential melanoma are much more severe than those of a false alarm.

\begin{table}[!t]
\caption{Five-fold evaluation results on the skin image dataset with different detectors. The inference time for each model is also reported with a single Nvidia Geforce RTX 2080Ti and 4 CPU cores. Models are pre-trained on COCO dataset and fine-tuned with the specifications provided in Subsection~\ref{sec:lesion-detector}.}
\centering
\resizebox{0.75\textwidth}{!}{%
\label{table:detection_results}
\begin{tabular}{lcc}
\toprule
Detector                                                    & mAP@IoU[0.5]  & Frame rate   \\ 
\midrule
YOLOv3 (adopted in~\cite{strzelecki2021skin})               & 63.3          & 128 \\
Faster RCNN (ResNet-50) (adopted in~\cite{zhao2021detection})& 84.6         & 17 \\
SDD (ResNet-18) + FPN (adopted in~\cite{mohseni2021can})    & 90.1          & 26 \\
ScaledYOLOv4 (our customised approach)                      & \textbf{92.4} & 47 \\
\bottomrule
\end{tabular}}
\end{table}

\begin{figure}[!t]
\begin{center}
\includegraphics[width=0.8\linewidth]{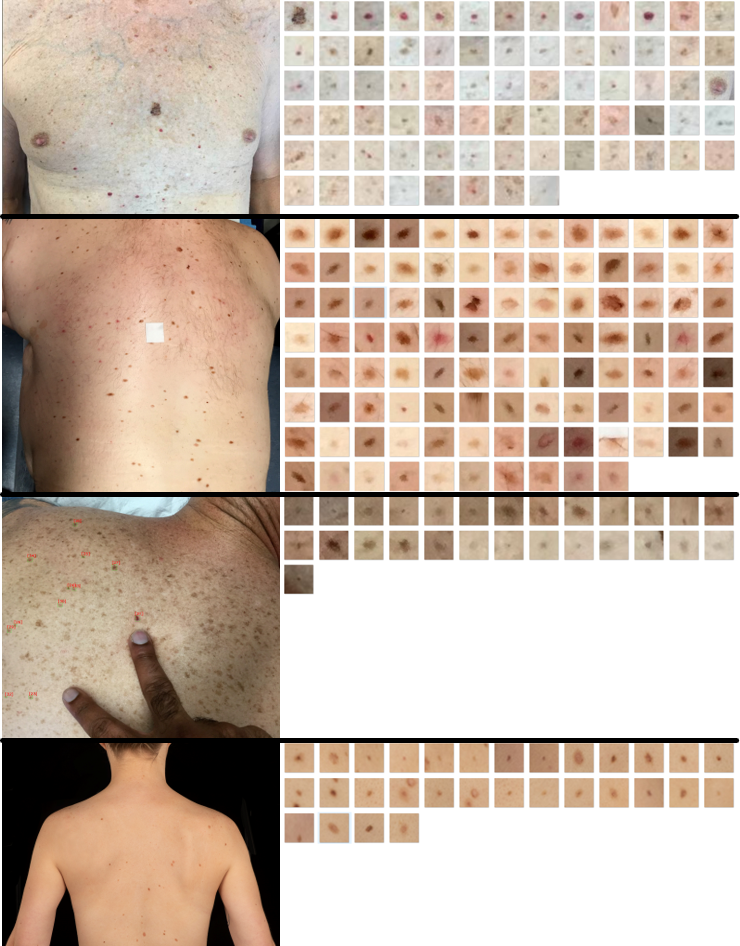}
\end{center}
\vspace{-14pt}
   \caption{Qualitative results of the skin lesion detector on clinical images introduced in Subsection~\ref{sec:dataset-skin}.
   }
\label{fig:qualitative_results_skin detector}
\end{figure}

The model, trained using the data specified in Table~\ref{table:dataset} achieved a mAP @ IoU[0.5] of 92.4\% with a precision of 79\% and a recall of 88\% on the test set (See Subsection~\ref{sec:lesion-detector}) while maintaining a fast frame rate of over 45 frames per second. This performance is comparable with the state-of-the-art methods for skin lesion detection on clinical images as shown in Table~\ref{table:detection_results}. 
Previous works on lesion detection or segmentation using high-magnification images of small regions surrounding a lesion obtained with a dermatoscope are not considered.
Our algorithm automatically detects skin lesions and provides an opportunity to perform quick screening of skin lesions in whole body images. This is especially useful for patients with many lesions.
Fig.~\ref{fig:qualitative_results_skin detector} shows some examples of the skin lesion detection in a visual format.

We also evaluated the performance of our detector on 55 distinct lesions identified from 3 participants (See Subsection~\ref{sec:data_collected}) and their various perspectives in all 2D images. This test aimed to assess the model's ability to detect the same lesion across different camera angles. The model achieved a mAP@IoU[0.5] of 85.1\%. However, the performance is degraded when the view of some lesions is distorted or blurred, as seen with Global ID 1 and its views from Poles C, N and M in Fig.~\ref{fig:global_ID}. It's important to note that these images were not used in the training of our detector.

\begin{figure}[!t]
\begin{center}
\includegraphics[width=0.70\linewidth]{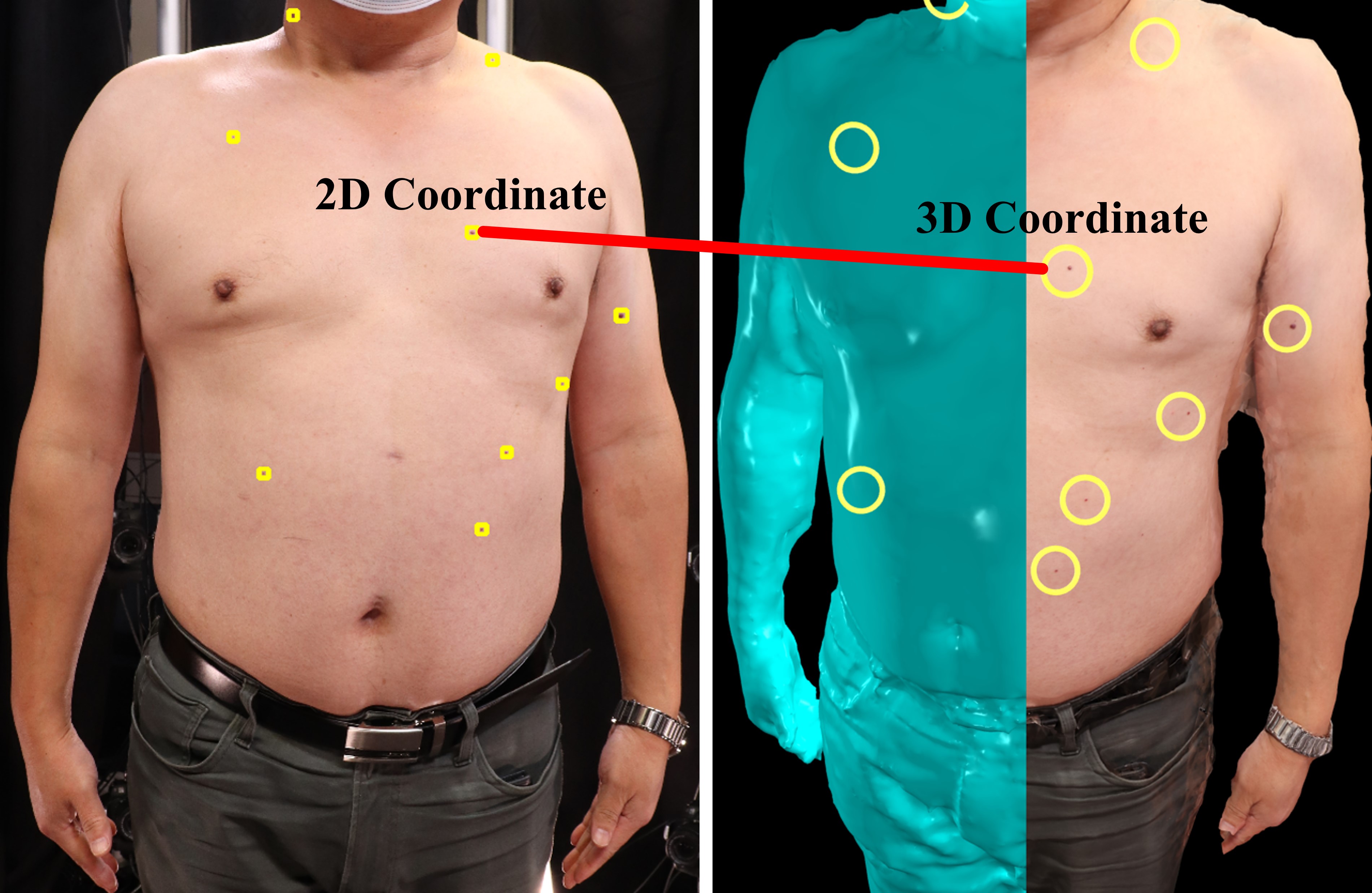}
\end{center}
\vspace{-9pt}
   \caption{Visualisation of the selected skin lesions detected in a selected 2D image [Front view of camera A3] (left) and mapped back to the 3D mesh of the subject [predefined front position in the interface] (right). To facilitate the visualisation of the 3D model, half of the human body is represented without texture. Since the camera poses and intrinsic parameters were estimated, it is possible to automatically map the 3D coordinates of a selected lesion to its corresponding 2D location on any of the regional images.
   }
\label{fig:qualitative_results_3D map}
\end{figure}

To overcome the limitation of identifying all visible lesions in all recorded images, we exploited 3D information by analysing the 3D positions of the lesions across multiple views. As described in Section~\ref{sec:2D-3D mapping lesion}, the system categorised the same skin lesion detected in multiple 2D images into a unique 3D global lesion identifier which is then used for temporal change analysis between sessions. This functionality enables the user to visualise the lesion from different angles, which can help overcome the limitations that may be present in a single image, such as a partial occlusion.
Fig.~\ref{fig:qualitative_results_3D map} shows the lesions detected on 2D images (frontal position), as well as the mapping of these lesions onto the 3D model reconstructed and visible in the user interface.
Additional qualitative results of the lesion detection on the collected data from a participant were shown in Fig.~\ref{fig:3D_collection} and Fig.~\ref{fig:qualitative_results_interface}.

\begin{figure}[!t]
\begin{center}
\includegraphics[width=0.98\linewidth]{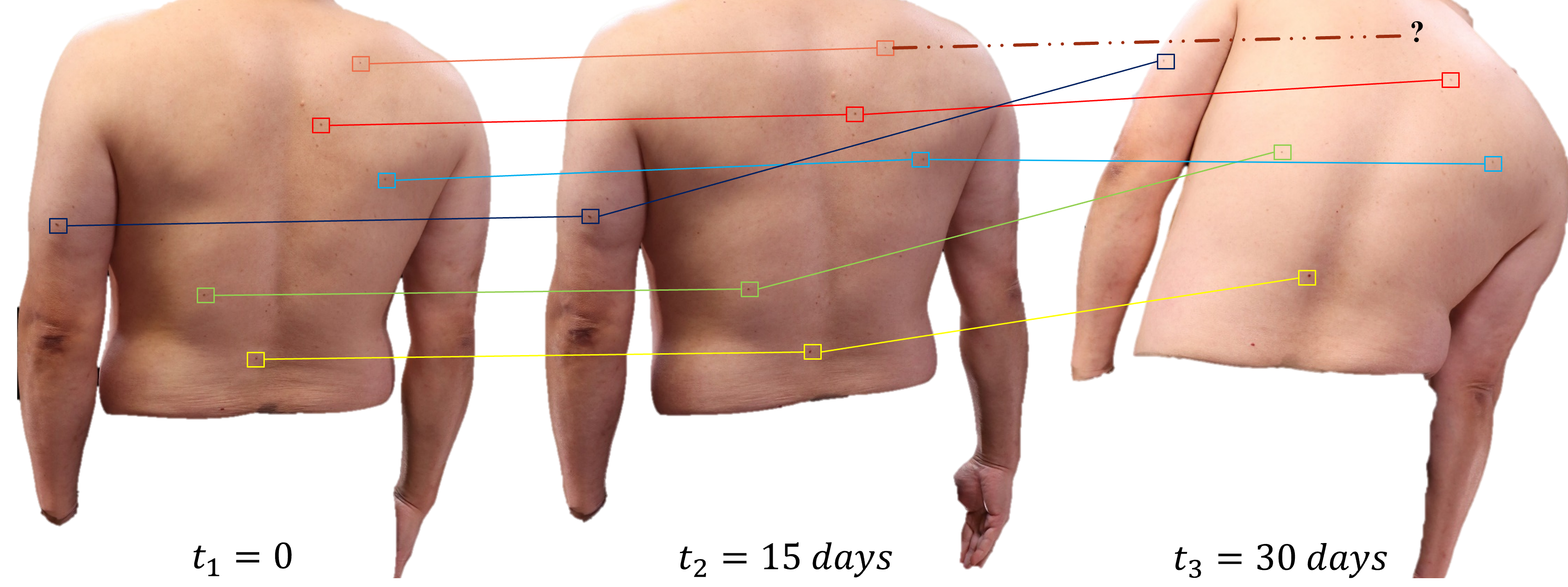}
\end{center}
\vspace{-13pt}
   \caption{Skin lesion matching. 3D meshes and longitudinal tracking of selected skin lesions across different sessions and poses ($t1=0$, $t2=15$ days, and $t3=30$ days). The coloured line between each lesion indicates the unique lesion correspondence across different meshes. The dotted line indicates missing lesion between $M_{t2}$ and $M_{t3}$.
   }
\label{fig:skin_tracking}
\vspace{-6pt}
\end{figure}

\subsubsection{Tracking of skin lesions across temporal scans}

To monitor the changes in lesions over time, we need to map the 2D locations of the lesions back to 3D coordinates with a unique ID. Then, we establish a match between the 3D human meshes and find the correspondence between the lesions on two consecutive meshes.
A metric called longitudinal accuracy is defined as the ratio of the number of lesion pairs that were correctly tracked to the total number of pairs within the longitudinal lesion set. It measures how accurately the lesions were tracked over time.
Using ten subjects scanned, three (3) from our image acquisition system and seven (7) from 3DBodyTexV1 (34 + 96 = 130 unique lesions) (Refer to Subsections~\ref{sec:data_collected}, ~\ref{sec:3D models}), we create a representative dataset to validate the matching across meshes.
While the lesion count is important to restrict the amount of required manual matching work, the diversity in body constitution and skin type help cover different skin topographies. Thus, with the different body characteristics in our experiments, the number of lesions is sufficient to evaluate the performance of the proposed method.

The solution based on the anatomical registration with LoopReg and the skin lesion matching with 3D geodesic distance reaches a longitudinal accuracy of 
94\% (32/34) between $M_{t1}$ and $M_{t2}$ (our dataset collected), 
88\% (85/96) between $M_{t1}$ and $M_{t2}$ (3DBodyTexV1 subjects), and
70\% (24/34) between $M_{t2}$ and $M_{t3}$ (our dataset collected). 
The accuracy of longitudinal examination may be impacted when it fails to track or match corresponding lesions if they are not detected by the lesion detector.
Fig.~\ref{fig:skin_tracking} illustrates the qualitative results of the longitudinal examination. While the majority of the observed lesions were successfully mapped, it can be seen that some lesions were not tracked due to detection failure on $t3=30$ days as a result of reflections in the image. Additionally, for this particular session, the examination faced challenges in tracking lesions that were heavily occluded on the right side of the body due to unusual body pose.


\section{Discussion}

3D whole body imaging is notably important in the analytical assessment of individual lesions relative to the overall skin lesion ecosystem of a patient~\cite{grochulska2021additive}.
Such technology is especially useful to track and document lesions of patients with many skin lesions, where diagnostic biopsy of every lesion is unfeasible. Our modular imaging system provides a promising tool to capture whole body images, detect and document lesions. 
This can potentially save large amounts of time for skin cancer specialists. The 3D imaging can be used to facilitate long-term surveillance by monitoring the entire skin surface rather than individual lesions~\cite{horsham2022experience}.

We have assessed the 3D reconstruction accuracy using synthetic images generated from existing 3D whole body scans, and discovered that the average error of the 3D reconstruction is about 2.5 mm. However, it is worth mentioning that the accuracy of the 3D reconstruction on the dataset collected with the proposed system was not studied explicitly, but rather the individual components of the system. While preferred, access to a certified commercial system such as VECTRA WB360 was not available to generate accurate ground truth data. Rotatory systems were not considered for this analysis.

While the primary aim of this study is to present a modular system that allows for 3D geometry analysis of skin lesions and data management, rather than a clinical study, we recognise that gathering additional data will facilitate appropriate validation of longitudinal tracking and patient variability for long-term surveillance.

If small lesions that occupy an area of less than $25 \times 25$ pixels (approximately $\leq$ 2 mm measured under real settings with the dataset collected with our proposed system) are to be detected, an unsatisfactory amount of false positives may be detected with the current quality of images. 
Earlier stages and smaller lesions of 1mm or 2mm in size, were much more difficult and error-prone~\cite{dildar2021skin}. To estimate the diameter of a lesion on these images, Otsu thresholding can be used to segment the lesion.
While nevi that are less than 2 mm in size may be less likely to be cancerous, it is still important to have them checked~\cite{koh2018mind}. 
Furthermore, for greater repeatability of 3D body imaging, only lesions $\geq 2$ mm have been considered for classification~\cite{betz2021reproducible,horsham2022experience}, as they are associated with a higher risk of cancer spreading (metastasis) or recurrence~\cite{NCCN}.
In general, smaller skin lesions are more likely to be confined to the top layer of the epidermis and less likely to have invaded deeper layers or spread to nearby lymph nodes~\cite{NCCN}.


Limitations of our model include the difficulties to filter the lesions based on more advanced geometrical parameters. 
By including a data-driven segmentation module, as shown in~\cite{mohseni2021can}, for each lesion within a bounding box, the system will be able to rank all the lesions with such characteristics.
Once the correspondence among overlapping images captured by adjacent cameras is established, the next steps are to segment the lesions and align the images to address the difference in perspectives, before identifying their temporal changes.
The accurate estimation of the diameter of a lesion is especially important when comparing the images captured over a period of time. 
A significant increase in the diameter of a nevus between consecutive examinations is a strong indicator that the melanocytic nevus can potentially be dangerous~\cite{koh2018mind,horsham2022experience}.
Additionally, our matching algorithm did not take into account the examination of meshes that showed lesions either disappearing or appearing anew~\cite{mirzaalian2016skin}.

\section{Conclusion}
In this study, we introduced a 3D whole body imaging system and workflow for multiple view image capturing, 3D human body reconstruction, 3D skin lesion localisation and documentation. A software tool has been designed for image visualisation and management. We have also developed an algorithm to identify skin lesions in whole body images. 
The baseline system that serves as a platform for new scientific discoveries is developed utilising open source libraries and software packages such as digiCamControl, Meshroom, Darknet, and QT via PySide2 for camera control, 3D human body reconstruction, lesion detection, and user interface development, respectively.
Future work will involve further evaluation of the workflow against a larger image set with expertly labelled lesions.

\bibliographystyle{splncs04}
\bibliography{egbib}
\end{document}